\documentclass[sigconf,nonacm]{acmart}
\AtBeginDocument{%
  }

\usepackage{multirow}
\usepackage{xcolor}
\usepackage{tcolorbox}
\tcbuselibrary{breakable}
\usepackage{fvextra}

\definecolor{bluebg}{HTML}{F0F8FF}
\definecolor{bluefg}{HTML}{0000FF}
\begin{document}

\title{Clarify Before Executing: A Self-Evolving Agent for Resolving Intent Asymmetry in 3D Tool Orchestration}

\author{Xiaoye Zhu}
\affiliation{%
	\institution{South China University of Technology}
	\city{Guangzhou}
	\state{Guangdong}
	\country{China}
}
\email{xyzhu1225@foxmail.com}

\author{Weixin Li}
\affiliation{%
	\institution{South China University of Technology}
	\city{Guangzhou}
	\state{Guangdong}
	\country{China}
}
\email{3536761163@qq.com}

\author{Junan Huo}
\affiliation{%
	\institution{South China University of Technology}
	\city{Guangzhou}
	\state{Guangdong}
	\country{China}
}
\email{huojunan0810@gmail.com}

\author{Bozhong Wang}
\affiliation{%
	\institution{South China University of Technology}
	\city{Guangzhou}
	\state{Guangdong}
	\country{China}
}
\email{1011361251@qq.com}

\author{Jia Zeng}
\affiliation{%
	\institution{South China University of Technology}
	\city{Guangzhou}
	\state{Guangdong}
	\country{China}
}
\email{1229239588@qq.com} 

\author{Yi Yang}
\affiliation{%
	\institution{South China University of Technology}
	\city{Guangzhou}
	\state{Guangdong}
	\country{China}
}
\email{ftyy@mail.scut.edu.cn}

\author{Cen Chen}
\affiliation{%
	\institution{South China University of Technology}
	\city{Guangzhou}
	\state{Guangdong}
	\country{China}
}
\email{chencen@scut.edu.cn}

\author{Qi Liu}
\correspondingauthor
\affiliation{%
	\institution{South China University of Technology}
	\city{Guangzhou}
	\state{Guangdong}
	\country{China}
}
\email{drliuqi@scut.edu.cn}

%% All authors above share the following affiliation.
%\affiliation{%
%	\institution{South China University of Technology}
%	\city{Guangzhou}
%	\state{Guangdong}
%	\country{China}
%}

\renewcommand{\shortauthors}{Zhu et al.}


\begin{abstract}
	A fundamental intent asymmetry plagues modern 3D asset creation: while state-of-the-art 3D toolchains demand precise, executable parameters, ordinary users typically provide vague, underspecified instructions. Current 3D agents treat this ambiguity as noise, defaulting to blind execution under a single-turn assumption. To address this limitation, we introduce \textsc{Clare}, a clarification-aware and evolutionary 3D agent that treats intent asymmetry not as an execution error, but as an opportunity for strategic dialogue. By decoupling the generation pipeline into four specialized cognitive roles, \textsc{Clare} intercepts and resolves underspecified instructions before invoking computationally expensive 3D tools to seamlessly execute tasks across five diverse domains: text-to-3D generation, single-view reconstruction, multi-view reconstruction, point cloud editing, and post-processing. Crucially, rather than relying on rigid manual rules, \textsc{Clare} self-evolves its clarification policy via simulated multi-turn interactions. By optimizing a Multi-turn Reward, the agent internalizes the delicate balance between interaction efficiency and task completion. To rigorously test this, we construct 3D-Clarify, a comprehensive benchmark comprising 620 interaction scenarios with systematically injected ambiguity, missing information, and mistaken details. \textsc{Clare} achieves state-of-the-art performance, with 60.40\% and 43.34\% success rates on single-step and multi-step tasks, respectively, more than doubling existing baselines. Both quantitative and qualitative results demonstrate that proactive clarification is the missing key to robust 3D execution.
	Code is available at \url{https://github.com/xyzhu1225/CLARE}.
\end{abstract}




\begin{teaserfigure}
	\includegraphics[width=\textwidth]{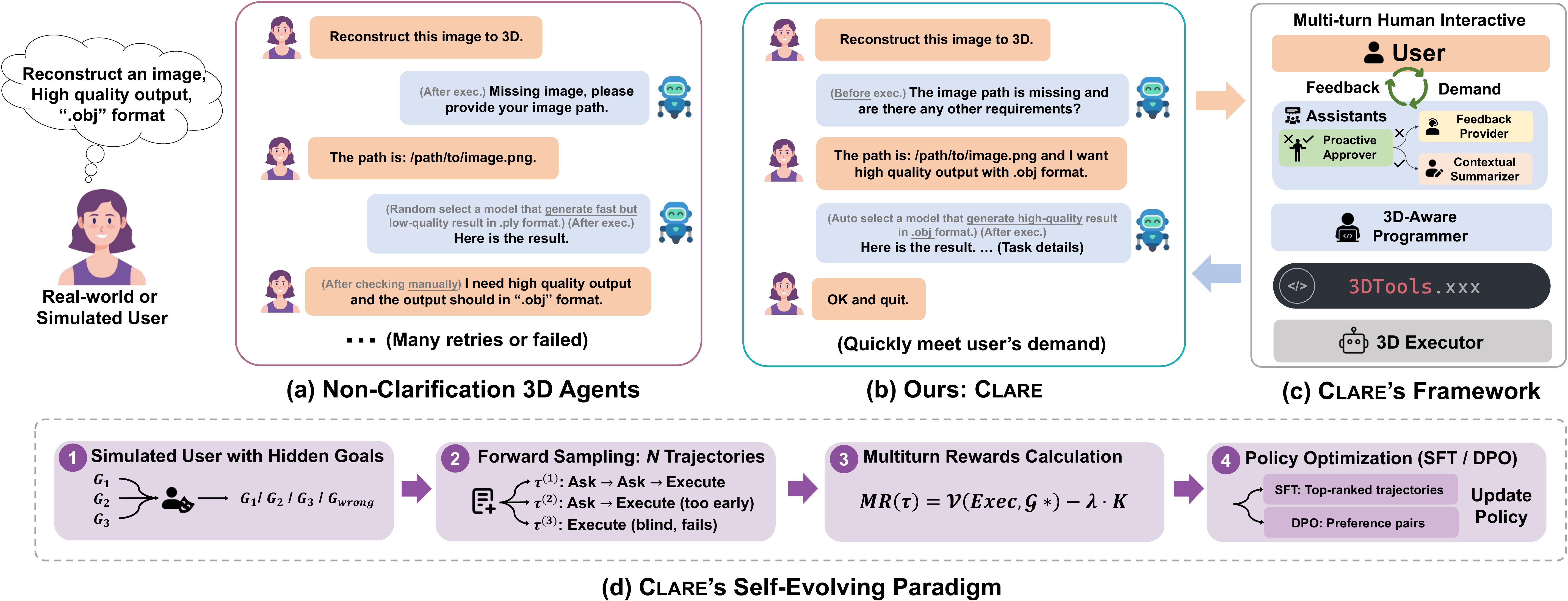}
	\caption{\textsc{Clare}'s framework and self-evolving pipeline.
		\textbf{(a-b)} Comparison: Traditional agents execute blindly vs. \textsc{Clare} proactively clarifies before execution.
		\textbf{(c)} Architecture: Modular components for approval, summarization, feedback, and 3D-aware programming.
		\textbf{(d)} Self-Evolving Pipeline: From sampling to fine-tuning.}
	\Description{The upper row contrasts a conventional agent that executes an underspecified 3D request before asking for missing details with CLARE, which clarifies first and routes the resolved request through an approver, summarizer, programmer, tool, and executor. The lower row shows self-evolution from hidden-goal simulated users through trajectory sampling and multi-turn reward calculation to SFT or DPO policy updates.}
	\label{fig:teaser}
\end{teaserfigure}

%\received[accepted]{9 July 2026}

\maketitle

\section{Introduction}

The democratization of 3D content creation has become increasingly critical across gaming, virtual reality, architectural visualization, and industrial design. Recent advances in 3D foundation models have enabled non-expert users to \textit{generate}, \textit{reconstruct}, and \textit{edit} 3D assets through natural language interfaces~\cite{lee2024text,jiang2024survey,zhang2025advances}. However, a fundamental intent asymmetry persists between \textit{what users naturally express} and \textit{what 3D systems strictly require}. While state-of-the-art 3D pipelines demand precise, executable parameters---such as explicit input file paths, strict output formats, appropriate model choices, and numerical quality constraints---ordinary users typically provide vague, underspecified instructions. A user might request "reconstruct this object" without providing the image, ask to "edit the chair" from a point cloud without specifying which points to keep, or inadvertently request incompatible parameters that lead to runtime failures.

Current 3D agent systems operate under a \textit{single-turn, well-specified assumption}: they expect complete and executable instructions in one shot. When faced with underspecified requirements, these systems exhibit three failure modes: (1) \textbf{blind execution} of ambiguous requests, producing irrelevant outputs that waste computational resources; (2) \textbf{undetected parameter contradictions}, causing cryptic runtime errors that frustrate users; and (3) \textbf{passive behavior}, lacking mechanisms to proactively identify and resolve information gaps. While general-purpose agentic frameworks like ReAct~\cite{yao2023react} and Reflexion~\cite{shinn2023reflexion} enable iterative reasoning in symbolic domains, they lack domain-specific mechanisms for detecting 3D-specific uncertainties---such as missing file paths, modality mismatches, or model capability violations.

Our key insight is that effective 3D task execution requires a \textit{progressive alignment} paradigm: \textbf{clarify before executing}. Rather than treating ambiguity as noise to be ignored, we posit that an intelligent 3D agent should (1) detect incomplete or mistaken requirements before invocation, (2) engage in targeted multi-turn clarification to resolve uncertainties, and (3) leverage domain knowledge to map user preferences (e.g., "high quality", "fast") to appropriate 3D models and parameters. Crucially, the decision of \textit{when to clarify} versus \textit{when to execute} is itself learnable—excessive clarification increases interaction cost, while premature execution risks failure.

To realize this vision, we introduce \textsc{Clare} (\textbf{Clar}ification-aware and \textbf{E}volutionary 3D agent), a multi-agent framework (Figure~\ref{fig:teaser}) that transforms underspecified user requirements into executable 3D workflows through structured collaboration. \textsc{Clare} decouples the generation pipeline into four specialized roles: (1) \textit{Proactive Approver}, which gates execution by detecting ambiguity, missing information, and mistaken details; (2) \textit{Contextual Summarizer}, which distills multi-turn dialogue into structured task specifications; (3) \textit{3D-Aware Programmer}, which synthesizes executable code by mapping user constraints to appropriate 3D tools; and (4) \textit{Feedback Provider}, which closes the interaction loop with human-readable status reports and targeted clarification questions. Beyond inference-time collaboration, \textsc{Clare} employs a \textit{self-evolution mechanism}: it learns optimal clarification policies through simulated interactions with a user agent that embodies realistic information asymmetry, optimizing a \textbf{Multi-turn Reward (MR)} that balances task completion against interaction efficiency. This simulation-based optimization enables the agent to internalize long-horizon decision making without manual annotation.

To rigorously evaluate \textsc{Clare}, we construct \textbf{3D-Clarify}, a comprehensive benchmark spanning five canonical 3D domains. Each task is systematically perturbed with three types of underspecified requirements---\textbf{ambiguity, missing information, and mistaken details}---creating a "lazy user" simulation that tests an agent's clarification capability. 

Our contributions are summarized as follows:
\begin{itemize}
    \item We propose \textsc{Clare}, the first multi-agent framework that systematically addresses underspecified user requirements in 3D task execution through proactive clarification and tool orchestration across five diverse domains: text-to-3D generation, single-view reconstruction, multi-view reconstruction, point cloud editing, and post-processing.
    \item We introduce 3D-Clarify, a benchmark with 620 tasks and three types of controlled underspecification. By standardizing the evaluation of intent asymmetry, it provides a foundational testbed to drive future research towards more robust, interactive 3D generation systems.
    \item We develop a self-evolution training paradigm that optimizes clarification policies via simulated user interactions and Multi-turn Reward (MR) optimization, enabling the agent to learn effective multi-turn decision making without manual annotation. This empowers the agent to dramatically elevate its execution success rate to 60.40\% on single-step tasks and 43.34\% on complex multi-step tasks, achieving robust performance with negligible interaction overhead.
\end{itemize}

\section{Related Work}

\subsection{LLM-based Agents for 3D Tasks}
Recent advances in large language models have enabled language-driven control over complex software systems, including 3D generation and reconstruction pipelines. In parallel, numerous agent reasoning frameworks have been employed to enable LLMs to interact with the external environment. For example, ReAct~\cite{yao2023react} interleaves reasoning traces with task-specific actions, allowing the agent to dynamically interact with the environment. Reflexion~\cite{shinn2023reflexion} further introduces a feedback-driven learning mechanism, enabling agents to improve subsequent decisions by reflecting on execution outcomes and past failures.

Existing LLM-based 3D agents~\cite{sun20243dgptprocedural3dmodeling,chen2024idea23dcollaborativelmmagents,yang2023llmgrounderopenvocabulary3dvisual,ai2025nli4volvisnaturallanguageinteraction} are mostly designed around specific 3D workflows rather than diverse task settings. By comparison, \textsc{Clare} targets a substantially broader task space, seamlessly integrating text-to-3D, reconstruction, point cloud editing, and post-processing. This capability to handle multi-step compositions makes our framework not only clarification-aware but also highly generalizable across unified 3D executions.

\subsection{Interactive Clarification}
Interactive clarification has been widely studied in human-centered AI systems~\cite{acikgoz2025speakrlsynergizingreasoningspeaking,acikgoz2025macmultiagentframeworkinteractive,qian2025userbenchinteractivegymenvironment,qian2025userrltraininginteractiveusercentric,sun2025trainingproactivepersonalizedllm}, where systems must actively gather missing information through multi-turn interactions before executing a task.

Existing work on interactive clarification can be broadly categorized into three directions. First, methods focusing on information elicitation ask follow-up questions to resolve underspecified constraints~\cite{zhang2024clamberbenchmarkidentifyingclarifying}. Second, user-centric frameworks explore interaction policy learning to balance task completion against dialogue cost~\cite{chen2024styleimprovingdomaintransferability,qian2025userrltraininginteractiveusercentric,kudrinskii2026faithful}. Third, collaborative paradigms shift from passive responses to active goal-oriented cooperation~\cite{zhang2024askbeforeplanproactivelanguageagents,huang2025teachinglanguagemodelsgather}. Collectively, these efforts establish clarification as a core decision-making capability.

Our work differs by situating clarification specifically within the fragile pipeline of 3D task execution. We treat dialogue not as an auxiliary behavior, but as a prerequisite to prevent invalid tool selections, parameter contradictions, and fatal downstream errors.

\subsection{Self-Evolving Agents}
Recent surveys~\cite{fang2025comprehensivesurveyselfevolvingai,gao2026a} suggest that the development of self-evolving agents reflects a paradigm shift from static systems to adaptive, lifelong agents that continuously improve through interaction.

However, self-evolving frameworks~\cite{zhou2024symboliclearningenablesselfevolving,yuan2025agentrtraininglanguagemodel} primarily focus on general language-centric reasoning, leaving domain-specific workflows like 3D asset creation underexplored. Our work fills this gap by introducing a 3D-domain-aware self-evolution paradigm. By coupling the agent with a stateful simulated user, we optimize clarification policies through multi-turn trajectories. This allows the system to internalize the delicate balance between interaction efficiency and reliable execution under underspecified requirements.

\section{Problem Formulation}

We formulate the 3D task execution problem as a goal-oriented interaction between a user and a 3D agent. Unlike traditional single-turn generation, the user possesses a \textit{latent goal} $\mathcal{G}^* = \{p_{in}, f, c, \theta\}$, which encompasses all essential information: the input source $p_{in}$ (path or prompt), the desired output format $f$, user preference constraints $c$ (e.g., quality, speed or other constraints), and task-specific parameters $\theta$. The initial user query $q_0$ is often an incomplete or noisy projection of $\mathcal{G}^*$. We formalize three types of \textit{underspecified requirements}:
\begin{itemize}
    \item \textbf{Ambiguity}: $q_0$ maps to a set of multiple candidate goals: $ q_0 \implies\{\mathcal{G}_1, \mathcal{G}_2, \dots, \mathcal{G}_n\}$.
    \item \textbf{Missing Info}: $q_0 \subset \mathcal{G}^*$, where one or more critical elements from $\{p_{in}, f, c, \theta\}$ are absent from $q_0$.
    \item \textbf{Mistaken Details}: $q_0$ contains explicit errors $e$ such that $e$ contradicts the physical constraints of the 3D environment or the logical consistency of $\mathcal{G}^*$. (e.g., offering wrong input files, out-of-bound parameters)
\end{itemize}

The interaction unfolds as a sequence of turns $T = \{(q_0, m_0), \dots,\allowbreak (q_K, m_K)\}$. At each turn $j$, the agent generates a response based on the updated history $\mathcal{H}_{<j}$: $\pi(q_j, \mathcal{H}_{<j}) \to (m_j, \mathcal{T}_{3D}, \theta_{exec})$, where $\mathcal{T}_{3D}$ indicates appropriate 3D tools and $\theta_{exec}$ corresponds to the execution parameters for solving the task. $(\mathcal{T}_{3D}, \theta_{exec}) = \emptyset$ if the agent chooses to clarify rather than execute. This loop continues until the user issues a ``Quit'' command or the interaction reaches the maximum turn limit $T_{max}$.  

Our objective is to learn an optimal policy $\pi^*$ that maximizes the expected completion rate of the 3D execution result while minimizing interaction cost:
\begin{equation}
    \pi^* = \arg\max_{\pi} \mathbb{E} \left[ \mathcal{V}(\text{Exec}(\mathcal{T}_{3D}, \theta_{exec}), \mathcal{G}^*) - \lambda \cdot K \right]
\end{equation}
where $\text{Exec}(\mathcal{T}_{3D}, \theta_{exec})$ denotes the output of the 3D toolchain, $\mathcal{V}(\cdot) \in [0, 1]$ is the value function provided by an automated verifier that assesses the final task completion rate, and $\lambda$ is a penalty for the total turns $K$.

\section{The CLARE Framework: Coordination via Progressive Alignment}

\subsection{Architecture Overview}
\begin{figure*}[htbp]
  \includegraphics[width=\textwidth]{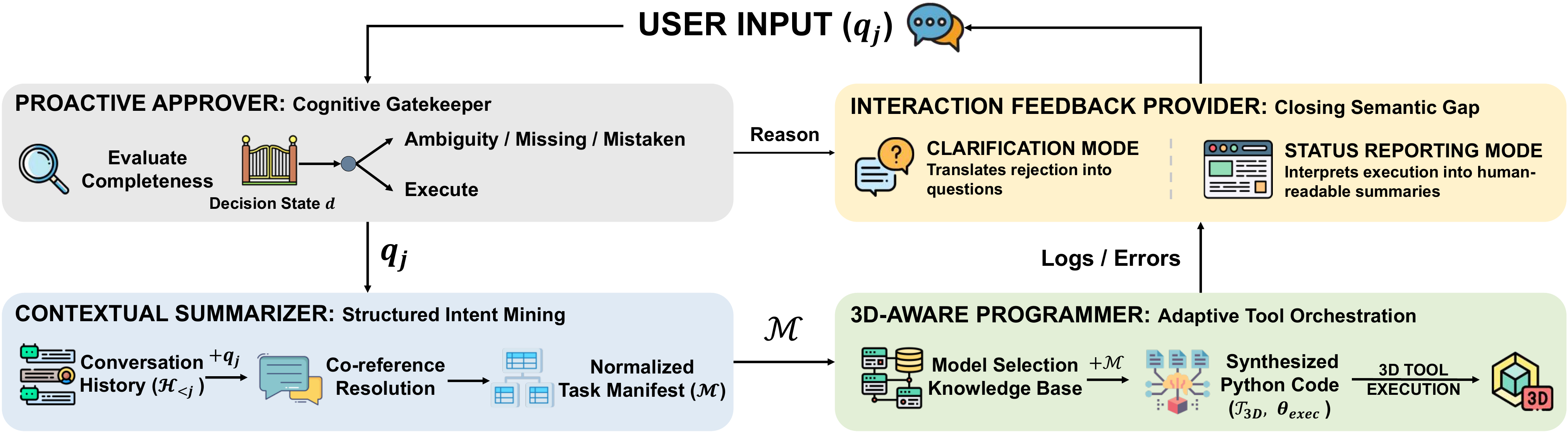}
  \caption{Architecture of \textsc{Clare}. Four specialized components collaborate to transform underspecified user queries into executable 3D workflows: Approver (gating), Summarizer (state tracking), Programmer (code synthesis), and Feedback Provider (conversational alignment). The system proactively clarifies underspecified requirements before execution.}
  \Description{A four-panel block diagram presents CLARE's components. The proactive approver chooses clarification or execution, the contextual summarizer converts dialogue into a normalized task manifest, the feedback provider asks questions or reports status, and the 3D-aware programmer selects tools and synthesizes executable Python code.}
  \label{fig:method}
\end{figure*}

Handling 3D tasks involves inherent ambiguity: users often lack the precise terminology to describe spatial geometry or rendering styles. We introduce \textsc{Clare}, a multi-agent framework designed not merely to execute commands, but to \textit{manage the uncertainty} of human intent. Figure~\ref{fig:method} shows a \textbf{progressive alignment loop} that decouples the complex generation process into four specialized cognitive roles: \textit{Decision Gating}, \textit{Context Refinement}, \textit{Executable Synthesis}, and \textit{Feedback Alignment}. This separation of concerns ensures that ambiguous or mistaken requirements are trapped and resolved \textit{before} touching the computationally expensive 3D tools.

\subsection{Core Components}

\paragraph{Proactive Approver: The Cognitive Gatekeeper}
Traditional agents often hallucinate parameters for vague instructions, leading to wasted generation time. The \textit{Proactive Approver} acts as the first line of defense. It evaluates the \textit{completeness} of the current instruction $q_j$ against the conversation history. Instead of blindly forwarding requests, it computes a binary decision state $d \in \{\textsc{Execute}, \textsc{Clarify}\}$. By detecting three distinct types of requirements—\textit{Ambiguity}, \textit{Missing Information}, and \textit{Mistaken Details}—the Approver effectively "freezes" the execution pipeline until the user's intent is fully actionable, enforcing a rigorous standard for task initiation.

\paragraph{Contextual Summarizer: Structured Intent Mining}
Once approved, the \textit{Contextual Summarizer} bridges the gap between fragmented natural language and structured execution specifications. It functions as the system's dynamic memory, aggregating the multi-turn dialogue history $\mathcal{H}_{<j}$ into a normalized task manifest $\mathcal{M}$. This component performs crucial \textit{state tracking}: it resolves co-references (e.g., "remove \textit{it}" referring to a previously generated chair) and explicitly extracts constraints such as target file paths and quality preferences. By distilling noise into a structured schema, it provides a noise-free specification for the downstream programmer.

\paragraph{3D-Aware Programmer: Adaptive Tool Orchestration}
Translating the Manifest $\mathcal{M}$ into code requires domain-specific knowledge. The \textit{3D-Aware Programmer} goes beyond generic code generation by incorporating a model selection knowledge base. It understands the capability boundaries of different 3D tools. As detailed in our implementation, the Programmer dynamically maps user-defined constraints $c$ (e.g., "high quality", "fast speed") to specific model $\mathcal{T}_{3D}$ and parameters $\theta_{exec}$. This ensures that the synthesized Python code is not only syntactically correct but physically and semantically grounded in the 3D generation context.

\paragraph{Interaction Feedback Provider: Closing the Semantic Gap}
Effective collaboration requires two-way communication. The \textit{Feedback Provider} serves as the interface for conversational alignment. It operates in two adaptive modes: \textbf{Clarification Mode:} When the Approver rejects a request, this module translates the rejection reasoning into a natural, targeted question, guiding the user to provide the missing bits without cognitive overload. \textbf{Status Reporting Mode:} Post-execution, it interprets raw execution logs and potential runtime errors, translating them into human-readable summaries. This closes the loop, allowing users to iteratively refine their 3D assets based on clear, actionable system feedback.

\subsection{Self-Evolution via Collaborative Simulation}
\label{sec:evolution}
Drawing inspiration from collaborative training frameworks \cite{wu2025collabllm}, we treat the 3D task execution as a long-term optimization problem. Static datasets often fail to capture the nuances of \textit{when} to clarify versus \textit{when} to execute. We address this by evolving the agent's policy $\pi$ through a \textbf{Collaborative Simulation Loop}, maximizing a multi-turn reward derived from interactive trajectories with a dynamic user simulator.

\subsubsection{Interactive Forward Sampling}  
To estimate the long-term utility of a clarification strategy, \textsc{Clare} performs lookahead via simulation. Given a context $x = (q_j, \mathcal{H}_{<j})$, the agent samples $N$ candidate responses $\{m^{(1)}, \dots, m^{(N)}\}$ from its current policy $\pi_\theta$. For each candidate, we roll out a full interaction trajectory $\tau^{(i)}$ by coupling the agent with a \textit{Simulated User} $\mathcal{U}$. This process reveals downstream consequences, helping the agent separate vital clarifications from redundant, costly queries.

\begin{figure*}[htbp]
	\includegraphics[width=\textwidth]{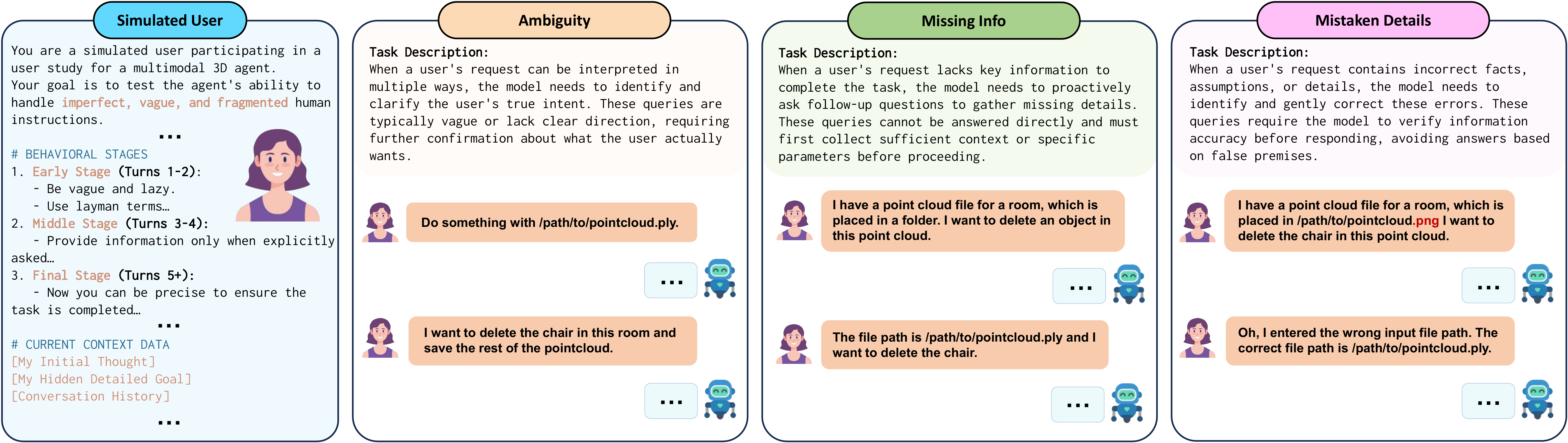}
	\caption{Overview of the 3D-Clarify benchmark. Each scenario starts with a fixed simulated user prompt, then presents an initial query with one of three types of underspecification: \textit{Ambiguity}, \textit{Missing Info}, or \textit{Mistaken Details}. The benchmark covers 620 interaction scenarios across five 3D domains. (500 single-step tasks and 120 multi-step tasks)}
	\Description{Four side-by-side panels show the simulated-user specification followed by benchmark examples for ambiguity, missing information, and mistaken details. Each category pairs its definition with a dialogue illustrating how hidden intent is revealed or corrected over multiple turns.}
	\label{fig:dataset}
\end{figure*}

\paragraph{Simulated User with Hidden Goals}  
The \textit{Simulated User} $\mathcal{U}$ is not a static responder but a stateful agent designed to mimic realistic human behavior. Initialized with a hidden target goal $\mathcal{G}^*$ (e.g., specific quality and format constraints) and a surface-level vague query $q_0$, $\mathcal{U}$ operates under an \textit{Information Asymmetry} setting. As detailed in our implementation, it follows strict behavioral stages: initially providing vague requirements and only revealing critical parameters when the agent explicitly identifies the information gap. This "interaction friction" forces the agent to learn precise interrogation strategies to unlock the hidden goal.
\subsubsection{Multi-turn Reward Formulation}  
We evaluate the quality of each sampled trajectory $\tau$ using a composite \textbf{Multi-turn Reward (MR)}, balancing task fidelity against interaction efficiency. The reward is computed by an automated \textit{Verifier}:
\begin{equation}
MR(\tau) = \mathcal{V}(\text{Exec}(\mathcal{T}_{3D}, \theta_{exec}), \mathcal{G}^*) - \lambda \cdot K
\end{equation}
where $K$ is the number of interaction turns, and $\lambda$ acts as a penalty coefficient to discourage verbosity. The verification function $\mathcal{V} \in [0, 1]$ assesses the semantic and parametric accuracy of the final 3D output:
\begin{equation}
\mathcal{V} = \mathbb{I}(\text{Type}_{\text{pred}} = \text{Type}_{\text{gt}}) \cdot \left[ \frac{1}{|\Theta|} \sum_{i=1}^{|\Theta|} \phi(\theta_i, \theta_i^*) \right]
\end{equation}
Here, $\mathbb{I}(\cdot)$ ensures the fundamental task type aligns. The second term computes the average parameter accuracy, where $|\Theta|$ denotes the total number of ground-truth parameters required by the target 3D tool. Each predicted parameter $\theta_i$ is scored against its ground-truth counterpart $\theta_i^*$ via the similarity function $\phi \in [0, 1]$, defined as:
% Here, $\mathbb{I}(\cdot)$ ensures the fundamental task type aligns. The second term computes the average precision of extracted parameters $\Theta$ (e.g., file paths, numerical constraints), where $\phi$ is a strict equality check for discrete values and a semantic similarity score valued by LLM for open-ended text descriptions. The similarity function $\phi(\theta_i, \theta_i^*) \in [0, 1]$ is defined as:
\begin{equation}
\phi(\theta_i, \theta_i^*) = 
\begin{cases} 
1 & \text{if } \theta_i = \theta_i^* \\
\text{LLM-Sim}(\theta_i, \theta_i^*) & \text{if } \theta_i, \theta_i^*~\text{are text} \\
0 & \text{otherwise}
\end{cases}
\end{equation}

\subsubsection{Policy Optimization}
To internalize the forward-looking capability derived from simulation, we optimize the agent's policy $\pi$ using the collected interaction trajectories. We investigate two distinct paradigms to refine the \textit{Approver}'s decision boundary: Supervised Fine-Tuning (SFT) and Direct Preference Optimization (DPO) \cite{rafailov2023direct}. Instead of relying on manual annotation, we leverage the automated metrics from our simulation loop to curate training signals. For SFT, we iteratively select
top-ranked responses to build realistic, goal-directed interaction trajectories. For DPO, we construct pairwise comparisons by ranking responses at each turn, distinguishing “chosen” and “rejected” pairs based on MR scores.

\section{The 3D-Clarify Benchmark}
\label{sec:dataset}

To rigorously evaluate \textsc{Clare}'s capability in ambiguity resolution and long-horizon coordination, we construct \textbf{3D-Clarify}, a comprehensive benchmark comprising 620 interaction scenarios. The construction pipeline is automated, consisting of three phases: atomic task generation, sequential composition, and uncertainty injection.

\subsection{Phase 1: Atomic Task Generation}
\begin{table}[htbp]
	\centering
	\caption{Single-turn atomic task distribution across 3D domains and sub-domains. The supported tools denote the current toolset integrated into our agent, while the dataset design ensures the integration of future tools.}
	\Description{The table groups 500 atomic tasks by five 3D domains and their sub-domains, listing the supported tools and the number of tasks in each category. Text-to-3D generation, single-view reconstruction, multi-view reconstruction, point-cloud editing, and post-processing each contribute 100 tasks.}
	\label{tab:task_dis}
	% 1. 缩短列间距（默认是6pt，改小至3pt或4pt）
	\setlength{\tabcolsep}{3pt} 
	% 2. 缩小整体字号以适应单栏
	\footnotesize 
	% 3. 第一列使用 p{宽度}，限制宽度并自动换行
	\begin{tabular}{p{1.8cm}lp{3.5cm}c}
		\toprule
		3D Task Domain & Sub-domain & Supported Tools & Count \\
		\midrule
		% 注意这里 \multirow 的第二个参数设定了与表头相同的固定宽度 2.8cm
		\multirow{2}{1.8cm}{Text-to-3D \\ Generation} & Scene & Worldgen~\cite{worldgen2025ziyangxie} & 40 \\
		& Object & Trellis~\cite{xiang2024structured}, Triplaneturbo~\cite{ma2025progressive} & 60 \\
		\midrule
		\multirow{3}{1.8cm}{Single-View \\ Reconstruction} & Scene & VGGT~\cite{wang2025vggt} & 20 \\
		& \multirow{2}{*}{Object} & VGGT~\cite{wang2025vggt}, Hunyuan-3D~\cite{lai2025hunyuan3d25highfidelity3d,hunyuan3d22025tencent,yang2024hunyuan3d} & \multirow{2}{*}{80} \\
		& & Trellis~\cite{xiang2024structured} & \\
		\midrule
		\multirow{2}{1.8cm}{Multi-View \\ Reconstruction} & Scene & VGGT~\cite{wang2025vggt}, Fast3R~\cite{yang2025fast3r} & 40 \\
		& Object & VGGT~\cite{wang2025vggt}, Fast3R~\cite{yang2025fast3r}, Trellis~\cite{xiang2024structured} & 60 \\
		\midrule
		\multirow{3}{1.8cm}{Point Cloud \\ Editing} & Keep & \multirow{3}{*}{Concerto~\cite{zhang2025concerto}} & 48 \\
		& Delete & & 48 \\
		& Segmentation & & 4 \\
		\midrule
		\multirow{2}{1.8cm}{Post-Processing} & Type Conversion & \multirow{2}{*}{Open3D~\cite{Zhou2018}} & 50 \\
		& Outlier Removal & & 50 \\
		\midrule
		Total & - & - & 500 \\
		\bottomrule
	\end{tabular}
\end{table}

As shown in Table~\ref{tab:task_dis}, we first establish a foundation of 500 single-turn atomic tasks. To ensure robust coverage, we define parameter spaces for five canonical 3D domains: \textit{Text-to-3D Generation}, \textit{Single-View Reconstruction}, \textit{Multi-View Reconstruction}, \textit{Point Cloud Editing}, and \textit{Post-Processing} (including Outlier Removal and Type Conversion).

For each task $\tau$, we employ a template-based generation engine. We map high-level user constraints (e.g., "high quality", "fast speed") to specific executable parameters $\Theta_\tau$ of compatible foundation models (e.g., \textit{Trellis}, \textit{Hunyuan3D}, \textit{VGGT}). By procedurally sampling valid parameter combinations and wrapping them in diverse natural language templates, we generate ground-truth triplets $(q, \mathcal{G}^*, \text{Code}^*)$. This ensures a rigorous mapping where every natural language instruction $q$ has a verifiable goal $\mathcal{G}^* = \{p_{in}, f, c, \theta\}$ and maps to a certain tool $\mathcal{T}_{3D}$ with required parameters $\theta_{exec}$.

Crucially, this construction methodology emphasizes \textbf{\textit{generalizability}}. By decoupling abstract user constraints from specific tool implementations, our benchmark is tool-agnostic: adapting to other 3D tools $\mathcal{T}_{3D}^*$ with similar functional properties requires only establishing a corresponding \textit{model mapping} (e.g., mapping "high quality" to a new model's specific high-resolution parameter) during evaluation. This design ensures that 3D-Clarify remains a relevant standard even as underlying foundation models upgrade.

\subsection{Phase 2: Sequential Multi-Task Composition}
Real-world 3D workflows often involve sequential dependencies. To simulate this, we construct 120 multi-step samples by chaining atomic tasks based on 15 distinct workflow templates (e.g., \textit{Reconstruction} $\rightarrow$ \textit{Editing} $\rightarrow$ \textit{Outlier Removal}). To mirror the natural ambiguity of human communication in sequential pipelines, we intentionally obfuscate the explicit file paths for intermediate inputs. Rather than providing absolute paths for downstream operations, the simulated user employs contextual co-references to denote intermediate results. This design explicitly challenges the agent to maintain a persistent state, requiring it to accurately resolve cross-turn dependencies and correctly map previous outputs to subsequent inputs to successfully execute the entire workflow.

\begin{table*}[htbp]
	\setlength{\tabcolsep}{3.0pt}
	\caption{Main results on the 3D-Clarify benchmark. We compare \textsc{Clare} against static single-turn baselines, dynamic agentic frameworks, and task-interface-matched clarification/planning/3D-agent adapters across three types of underspecified requirements. We report Goal Completion Rate (CR\%) and Task Success Rate (SR\%) for both single-step (500 tasks) and multi-step (120 tasks) settings. ``SE'' means self-evolution via MR optimization.}
	\Description{Rows compare static prompts, dynamic agentic baselines, related-work adapters, and three CLARE variants; columns report completion and success rates for ambiguity, missing-information, mistaken-detail, and average cases on single-step and multi-step tasks. CLARE with SE-DPO achieves the highest average success rate in both settings, 60.40 percent and 43.34 percent, respectively.}
	\label{tab:main}
	\begin{tabular}{l|cccccccc|cccccccc}
		\toprule
		\multirow{3}{*}{\textbf{Method}} & \multicolumn{8}{c|}{\textbf{500 Single-step Tasks}} & \multicolumn{8}{c}{\textbf{120 Multi-step Tasks}} \\ \cmidrule(lr){2-9}  \cmidrule(lr){10-17} 
		\multicolumn{1}{c|}{} & \multicolumn{2}{c}{Ambiguity} & \multicolumn{2}{c}{Missing} & \multicolumn{2}{c|}{Mistaken} & \multicolumn{2}{c|}{Average} & \multicolumn{2}{c}{Ambiguity} & \multicolumn{2}{c}{Missing} & \multicolumn{2}{c|}{Mistaken} & \multicolumn{2}{c}{Average} \\ \cmidrule(lr){2-9}  \cmidrule(lr){10-17} 
		\multicolumn{1}{c|}{} & \%CR & \%SR & \%CR & \%SR & \%CR & \multicolumn{1}{c|}{\%SR} & \%CR & \%SR & \%CR & \%SR & \%CR & \%SR & \%CR & \multicolumn{1}{c|}{\%SR} & \%CR & \%SR \\ \midrule
		Zeroshot & 31.90 & 4.00 & 17.25 & 0.00 & 77.55 & \multicolumn{1}{c|}{34.80} & 42.23 & 12.93 & 20.43 & 0.00 & 21.62 & 0.00 & 85.31 & \multicolumn{1}{c|}{35.83} & 42.45 & 11.94 \\
		Fewshot & 34.92 & 6.00 & 18.42 & 0.00 & 77.25 & \multicolumn{1}{c|}{32.60} & 43.53 & 12.87 & 23.03 & 0.00 & 20.86 & 0.00 & \textbf{89.04} & \multicolumn{1}{c|}{41.67} & 44.31 & 13.89 \\
		CoT \cite{wei2022chain} & 31.38 & 4.40 & 18.15 & 0.00 & 73.03 & \multicolumn{1}{c|}{25.80} & 40.85 & 10.07 & 19.92 & 0.00 & 20.88 & 0.00 & \underline{86.17} & \multicolumn{1}{c|}{34.17} & 42.32 & 11.39 \\
		ReAct \cite{yao2023react} & 29.88 & 5.40 & 22.42 & 1.60 & 50.38 & \multicolumn{1}{c|}{16.40} & 34.23 & 7.80 & 14.51 & 0.00 & 18.04 & 0.00 & 74.72 & \multicolumn{1}{c|}{13.33} & 35.76 & 4.44 \\
		Reflexion \cite{shinn2023reflexion} & 39.47 & 8.40 & 28.07 & 2.40 & 69.31 & \multicolumn{1}{c|}{37.60} & 45.62 & 16.13 & 22.53 & 0.00 & 25.60 & 0.00 & 83.68 & \multicolumn{1}{c|}{23.33} & 43.94 & 7.78 \\
		\midrule
		CLAMBER \cite{zhang2024clamberbenchmarkidentifyingclarifying} & 50.87 & 26.80 & 44.85 & 13.80 & 71.32 & \multicolumn{1}{c|}{32.20} & 55.68 & 24.27 & 13.12 & 2.50 & 21.23 & 4.17 & 77.12 & \multicolumn{1}{c|}{21.67} & 37.16 & 9.44 \\
		CEP \cite{zhang2024askbeforeplanproactivelanguageagents} & 57.94 & 37.60 & 41.48 & 19.20 & 36.52 & \multicolumn{1}{c|}{16.40} & 45.31 & 24.40 & 25.20 & 5.00 & 21.87 & 5.83 & 48.70 & \multicolumn{1}{c|}{8.33} & 31.92 & 6.39 \\
		3D-GPT \cite{sun20243dgptprocedural3dmodeling} & 35.18 & 10.60 & 34.41 & 10.60 & 56.04 & \multicolumn{1}{c|}{25.20} & 41.88 & 15.47 & 11.18 & 2.50 & 21.24 & 1.67 & 68.85 & \multicolumn{1}{c|}{20.00} & 33.76 & 8.06 \\
		\midrule
		\textsc{Clare-base} & 59.31 & 45.80 & 57.58 & 41.00 & \underline{81.90} & \multicolumn{1}{c|}{59.00} & 66.26 & 48.60 & 38.16 & \underline{23.33} & 40.73 & 22.5 & 81.25 & \multicolumn{1}{c|}{35.83} & 53.38 & 27.22 \\
		\quad \textsc{$-$ SE-SFT} & \underline{62.29} & \underline{49.00} & \underline{69.47} & \underline{52.00} & \textbf{85.43} & \multicolumn{1}{c|}{\textbf{66.80}} & \underline{72.40} & \underline{55.93} & \underline{49.72} & \textbf{34.17} & \underline{68.10} & \underline{44.17} & 83.77 & \multicolumn{1}{c|}{\textbf{48.33}} & \textbf{67.20} & \underline{42.22} \\
		\quad \textsc{$-$ SE-DPO} & \textbf{70.52} & \textbf{60.00} & \textbf{75.32} & \textbf{59.40} & 74.68 & \multicolumn{1}{c|}{\underline{61.80}} & \textbf{73.51} & \textbf{60.40} & \textbf{53.41} & \textbf{34.17} & \textbf{68.74} & \textbf{49.17} & 72.75 & \multicolumn{1}{c|}{\underline{46.67}} & \underline{64.97} & \textbf{43.34} \\ \bottomrule
	\end{tabular}
\end{table*}

\subsection{Phase 3: Synthesizing Underspecified Requirements}
\label{sec:imperfect_synthesis}
To systematically evaluate clarification capabilities, we apply three controlled perturbation strategies to the ground-truth goals (Figure~\ref{fig:dataset}) to simulate a "Lazy User" persona. Building upon our problem formulation, we synthesize: (1) Ambiguity Injection, where we apply semantic bleaching via task-specific templates, replacing domain-specific terms with generic actions and stripping parametric constraints; (2) Missing Information, where we mask concrete execution parameters with indefinite placeholders while retaining structural intent; and (3) Mistaken Details, which simulate erroneous user beliefs requiring active correction. This is achieved by injecting contradictions, input modality mismatches (e.g., providing a 2D .jpg for a task requiring a 3D .ply), and out-of-bound values.

\section{Experiments}

\subsection{Experiment Setup}
To rigorously evaluate the capability of agents in handling realistic user requirements, we utilize our proposed \textbf{3D-Clarify} benchmark, which comprises 500 single-step and 120 multi-step tasks, each associated with three underspecified queries. Diverging from single-turn paradigms, we emphasize dynamic goal fulfillment by leveraging Qwen3-8B as a stateful user simulator to emulate authentic, multi-turn behaviors conditioned on the initial query, final objective, and dialogue history. Our agent components are similarly powered by Qwen3-8B. 

To enable adaptive tool orchestration, we construct a comprehensive model knowledge base by analyzing a diverse suite of integrated 3D toolsets, including Worldgen~\cite{worldgen2025ziyangxie}, Trellis~\cite{xiang2024structured}, Triplaneturbo~\cite{ma2025progressive}, Hunyuan-3D~\cite{lai2025hunyuan3d25highfidelity3d,hunyuan3d22025tencent,yang2024hunyuan3d}, VGGT~\cite{wang2025vggt}, Fast3R~\cite{yang2025fast3r}, Concerto~\cite{zhang2025concerto}, and Open3D~\cite{Zhou2018}. For self-evolution, we perform LoRA-based fine-tuning on the \textit{Proactive Approver} using a de-duplicated subset of 3,000 multi-turn trajectories filtered by the Multi-turn Reward. Both Supervised Fine-Tuning (SFT) and Direct Preference Optimization (DPO) are conducted using the LLaMA-Factory~\cite{zheng2024llamafactory} framework on two NVIDIA RTX 3090 GPUs for 2 epochs, utilizing a learning rate of $5 \times 10^{-6}$ and a total batch size of 4. For a fair comparison, all baselines employ the identical Qwen3-8B backbone. For \textsc{Clare} and the original baselines, inference uses temperature $0.6$ and top\_p $0.9$; the structured-output adapters use temperature $0.2$ and top\_p $0.9$, as detailed in Appendix~\ref{app:training-details}. All interactive methods use a maximum interaction limit of $T_{max}=8$ turns. In the reward formulation, the penalty coefficient $\lambda$ is set to $0.1$.

Detailed agent prompts, tool specifications, benchmark construction, and model-selection knowledge are provided in Appendices~\ref{app:method-details} and~\ref{app:benchmark-details}.

\subsection{Baselines}
To evaluate the effectiveness of our proposed framework, we compare against static single-turn prompting methods (Zero-shot, Few-shot, and Chain-of-Thought~\cite{wei2022chain}) and dynamic, feedback-driven agentic frameworks (ReAct~\cite{yao2023react} and Reflexion~\cite{shinn2023reflexion}). We further include three task-interface-matched related-work adapters: CLAMBER~\cite{zhang2024clamberbenchmarkidentifyingclarifying}, which uses a single clarification-aware agent; CEP~\cite{zhang2024askbeforeplanproactivelanguageagents}, which follows an ask-before-plan state collection, clarification, and planning flow; and 3D-GPT~\cite{sun20243dgptprocedural3dmodeling}, which follows task dispatch, conceptualization, and modeling. These adapters retain their defining control flows while using the same Qwen3-8B backbone, simulated user, 3D tool interface, turn budget, and evaluator as \textsc{Clare}; they are adaptations to our executor rather than exact reproductions. Full implementation details are provided in Appendix~\ref{app:baseline-details}.
% To evaluate the effectiveness of our proposed framework, we compare our method against two categories of representative baselines:
% \paragraph{Static LLM-based Approaches} These works focus on the model's capability to execute 3D tasks in a single turn without proactive clarification. \textbf{Zero-shot}: The agent directly generates code based on the user's initial (potentially underspecified) query without any prior examples or reasoning steps. \textbf{Few-shot}: The agent is provided with several in-context examples of successful 3D task executions to guide its code generation. \textbf{Chain-of-Thought (CoT)}~\cite{wei2022chain}: We prompt the model to generate a step-by-step plan (reasoning chain) before producing the final executable code, enhancing its logic in complex scene manipulation.

% \paragraph{Dynamic Agent-based Approaches}
% These methods incorporate iterative loops or feedback mechanisms to handle complex or underspecified instructions. \textbf{ReAct}~\cite{yao2023react}: A paradigm that combines reasoning and acting. The agent generates reasoning traces and task-specific actions in an interleaved manner, allowing it to interact with the environment. \textbf{Reflexion}~\cite{shinn2023reflexion}: An advanced agentic framework that reinforces the model through linguistic feedback. It attempts to self-correct previous failures by reflecting on execution errors or user dissatisfaction.

\subsection{Evaluation Metrics}
We adopt two widely recognized metrics to evaluate agent performance on the benchmark: Goal Completion Rate (CR), which quantifies the extent of task fulfillment for each individual task, and Task Success Rate (SR), which measures the ratio of tasks that are fully accomplished. For each task, we verify:
(1) \textit{Module Name}: correctness of the target module identification;
(2) \textit{Prompt \& Parameters}: consistency of text-to-3D prompts validated by LLM judgment, along with other generation parameters;
(3) \textit{Input File Path}: existence and validity of referenced input assets;
(4) \textit{Output File Extension}: correctness of naming conventions (e.g., .glb, .obj).

\subsection{Quantitative Results}
\begin{figure*}[htbp]
	\includegraphics[width=\textwidth]{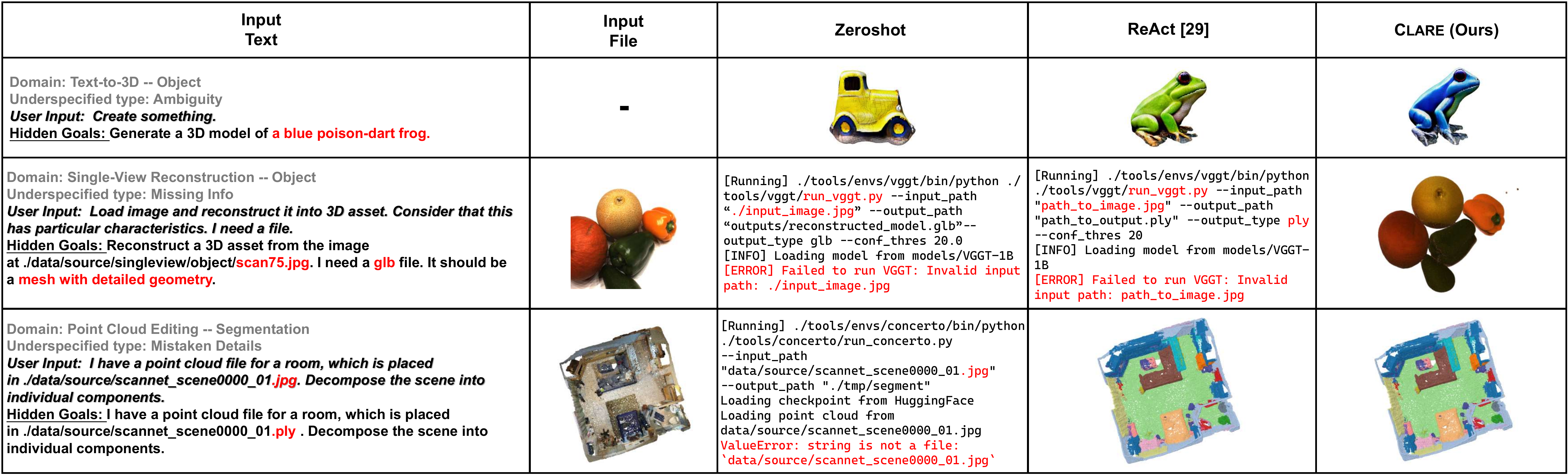}
	\caption{Qualitative comparison of 3D task execution under underspecified user requirements. While traditional baselines default to blind execution---resulting in hallucinated geometries or fatal execution errors due to ungrounded paths and modality conflicts---our agent proactively clarifies user intent to generate accurate, high-quality 3D assets.}
	\Description{Three rows compare Zero-shot, ReAct, and CLARE on text-to-3D ambiguity, reconstruction with missing input information, and point-cloud editing with an incorrect file extension. The baseline columns contain an incorrect object or invalid-path failures, whereas CLARE consistently produces the intended blue frog, reconstructed fruit mesh, and segmented room point cloud.}
	\label{fig:qualitative}
\end{figure*}
Table~\ref{tab:main} presents the comparison on the 3D-Clarify benchmark. We emphasize \textbf{Task Success Rate (SR)} as the primary metric for flawless execution, while \textbf{Goal Completion Rate (CR)} reflects partial parametric accuracy. Our analysis yields three key findings.

\paragraph{Proactive Clarification is Essential for Underspecified Requirements.}
For single-step tasks, single-turn paradigms (Zero-shot, Few-shot, CoT) nearly collapse on ambiguity and missing-info tasks, achieving only 0--6\% SR despite moderate CR scores (31--35\% and 17--18\%, respectively). Dynamic agentic frameworks (ReAct, Reflexion) show similar failures (1.6--8.4\% SR on ambiguity/missing tasks). Adding clarification-aware or 3D-specific scaffolds improves some cases, but the best adapter reaches only 24.40\% average SR on single-step tasks and 9.44\% on multi-step tasks. In contrast, \textsc{Clare-base} achieves 48.60\% and 27.22\% average SR, respectively. This gap indicates that an ask-first prompt or standard 3D decomposition alone is insufficient; reliable execution also requires explicit pre-execution gating and persistent clarified state.

\paragraph{Self-Evolution via MR Optimization Substantially Improves Clarification Policies.}
Multi-turn reward optimization yields significant gains in both SR and CR. On ambiguity single-step tasks, \textsc{Clare-SE-DPO} achieves 60.00\% SR (CR: 70.52\%) compared to \textsc{Clare-base}'s 45.80\% SR---a 14.20\% absolute improvement. On missing-info tasks, \textsc{Clare-SE-SFT} reaches 52.00\% SR (CR: 69.47\%), while \textsc{Clare-SE-DPO} achieves 59.40\% SR (CR: 75.32\%). For mistaken tasks, \textsc{Clare-SE-SFT} attains the highest 66.80\% SR (CR: 85.43\%). Averaged across all single-step underspecifications, \textsc{Clare-SE-DPO} achieves 60.40\% SR (CR: 73.51\%), more than doubling the strongest baseline (CEP: 24.40\% SR) and demonstrating that clarification policies are learnable through simulated interaction.

\paragraph{Multi-Step Tasks Amplify the Advantage of Structured Clarification.}
The transition to multi-step workflows reveals a critical finding: baseline methods collapse to 0\% SR on ambiguity and missing-info multi-step tasks, while \textsc{Clare-SE-DPO} maintains 34.17--49.17\% SR (CR: 53.41--68.74\%). This 34--49\% absolute gap underscores that multi-step workflows require \textit{consistent} clarification across turns---a capability enabled by \textsc{Clare}'s \textit{Contextual Summarizer} for maintaining structured task state. On mistaken multi-step tasks, \textsc{Clare-SE-SFT} achieves 48.33\% SR (CR: 83.77\%), demonstrating that early-stage clarification prevents error propagation. Overall, \textsc{Clare-SE-DPO} attains 43.34\% average SR (CR: 64.97\%) on multi-step tasks, compared to 4.44--7.78\% SR for dynamic baselines and 11.39--13.89\% SR for single-turn paradigms. While absolute performance indicates room for growth (e.g., larger backbones), \textsc{Clare} represents the first demonstration that \textit{proactive clarification} is the key to reliable 3D task execution under underspecified requirements. 

\paragraph{Performance Analysis on Mistaken Details.} We observe a striking divergence between Goal Completion Rate (CR) and Task Success Rate (SR) across baselines on mistaken tasks. For instance, single-step Zero-shot achieves 77.55\% CR but only 34.80\% SR. Because the underlying LLM possesses strong instruction-following capabilities, baselines successfully configure most standard parameters, artificially inflating CR. However, lacking proactive clarification, they blindly execute intentionally injected fatal errors, causing the overall execution to collapse. Conversely, our framework intercepts these errors, though CLARE-SE-SFT outperforms CLARE-SE-DPO (66.80\% vs. 61.80\% single-step SR). We attribute this to the inherent conflict between preference alignment and error correction. SFT directly clones robust "firm correction" trajectories via maximum likelihood estimation. In contrast, preference optimization (DPO) biases the model toward compliant, agreeable responses. Consequently, when confronted with explicit user errors, the DPO agent experiences policy degradation in the challenging action of proactively contradicting the user, leading to lower success rates.

\subsection{Qualitative Evaluation of 3D Execution}
Figure~\ref{fig:qualitative} visualizes \textsc{Clare}'s robustness across three underspecification scenarios:
\textbf{Overcoming Blind Execution}: Baselines hallucinate parameters for ambiguous requests (e.g., a generic frog), whereas CLARE clarifies the target ("blue poison-dart frog").
\textbf{Bridging Information Gaps}: Baselines fatally crash by executing placeholder code or selecting default tools. CLARE proactively queries missing parameters (e.g., image paths) prior to invocation.
\textbf{Preempting Failures}: Baselines pass invalid parameters causing API crashes. CLARE identifies constraint violations pre-execution, confirming proactive gating outperforms post-hoc recovery.

\subsection{Interaction Efficiency Analysis}
\label{sec:efficiency}

\begin{figure}[htbp]
	\centering
	\includegraphics[width=\linewidth]{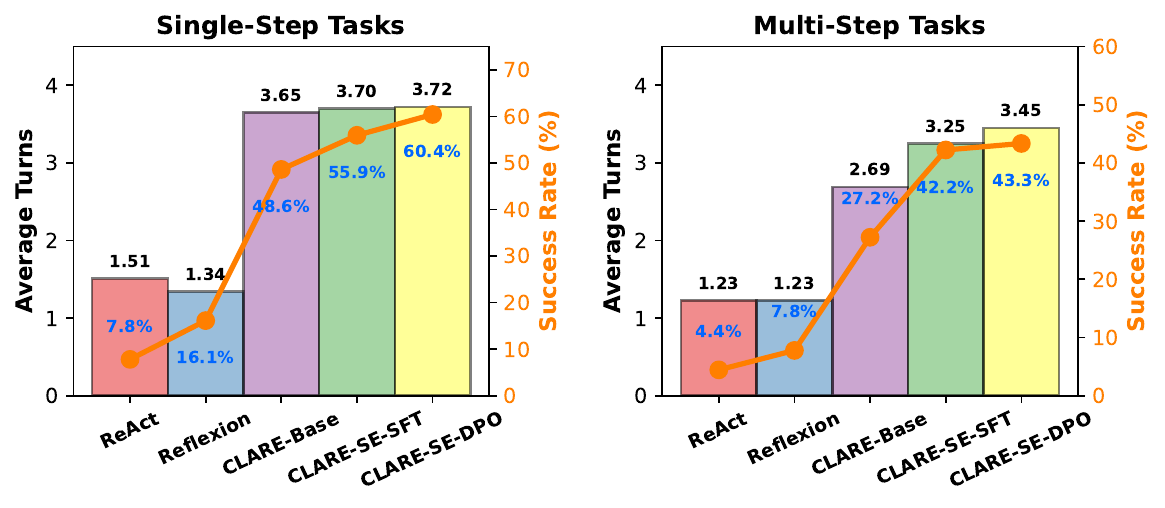}
	\caption{Comparison of average interaction turns and success rates across methods.
		\textbf{Left}: Single-step tasks. \textbf{Right}: Multi-step tasks.
		Multi-turn Reward training achieving the best trade-off between interaction cost and task completion.}
	\Description{Two dual-axis plots compare five methods on single-step and multi-step tasks, using bars for average interaction turns and orange lines for success rate. CLARE variants use roughly 2.7--3.7 turns and achieve 27.2--60.4 percent success, substantially above ReAct and Reflexion at about 1.2--1.5 turns and 4.4--16.1 percent, with DPO highest in both settings.}
	\label{fig:turns}
\end{figure}

Figure~\ref{fig:turns} illustrates the trade-off between interaction cost and task performance. ReAct and Reflexion require only 1.34--1.51 turns on single-step tasks, but achieve SRs of merely 7.8\% and 16.1\%, respectively. In contrast, \textsc{Clare-base} uses 3.65 turns to reach 48.6\% SR on single-step tasks and 2.69 turns to reach 27.2\% SR on multi-step tasks. MR optimization adds only 0.07 and 0.76 turns for \textsc{Clare-SE-DPO}, while increasing SR to 60.4\% and 43.3\%, respectively. These results show that strategic clarification substantially improves task completion at a modest interaction cost.

\subsection{Ablation Study}

To quantify the contribution of individual modules within \textsc{Clare}, we conduct ablation studies by disabling the \textit{Proactive Approver} and the \textit{Contextual Summarizer}. Table~\ref{tab:ablation} reports the results.
\begin{table}[htbp]
	\centering
	\caption{Ablation study on 3D-Clarify's single-step and multi-step tasks. Results are averaged across three types of underspecification (Ambiguity, Missing, Mistaken).}
	\Description{Rows compare CLARE-base with variants that remove the Summarizer or Approver, and columns report completion and success rates on single-step and multi-step tasks. Removing the Approver causes the largest decline, reducing success rate from 48.60 to 27.67 on single-step tasks and from 27.22 to 13.06 on multi-step tasks.}
	\label{tab:ablation}
	\begin{tabular}{lcccc}
		\toprule
		\multirow{2}{*}{\textbf{Method}} & \multicolumn{2}{c}{\textbf{Single}} & \multicolumn{2}{c}{\textbf{Multi}} \\
		\cmidrule(lr){2-3} \cmidrule(lr){4-5}
		& \%CR & \%SR & \%CR & \%SR \\
		\midrule
		\textsc{Clare-base} & \textbf{66.26} & \textbf{48.60} & \textbf{53.38} & \textbf{27.22} \\
		\quad $-$ w/o \textit{Summarizer} & \underline{65.25} & \underline{48.20} & \underline{45.97} & \underline{20.28} \\
		\quad $-$ w/o \textit{Approver}   & 52.04 & 27.67 & 38.19 & 13.06 \\
		\bottomrule
	\end{tabular}
\end{table}

Removing the Approver causes the most severe performance degradation: on single-step tasks, SR drops from 48.60\% to \textbf{27.67\%} (\textbf{-20.93\%} absolute), and CR falls from 66.26\% to 52.04\% (-14.22\%). On multi-step tasks, the SR decline is also substantial (from 27.22\% to 13.06\%, -14.16\%). This validates our central hypothesis: the ability to detect ambiguity, missing information, and mistaken details \textit{before} execution is the foundation of \textsc{Clare}'s effectiveness. Without this gating mechanism, the system reverts to blind execution, hallucinating parameters and failing to resolve information gaps.

The Summarizer exhibits complexity-dependent utility. While its impact on single-step tasks is negligible ($\Delta$SR: -0.40\%), its absence significantly impairs multi-step workflows, where SR drops by 6.94\% and CR by 7.41\%. This divergence validates our design rationale: while the Summarizer is redundant for atomic instructions, it is indispensable for maintaining structured state and resolving cross-turn dependencies as interaction depth increases.

\subsection{Effect of Training Data Scale}

\begin{figure}[htbp]
	\centering
	\includegraphics[width=0.8\linewidth]{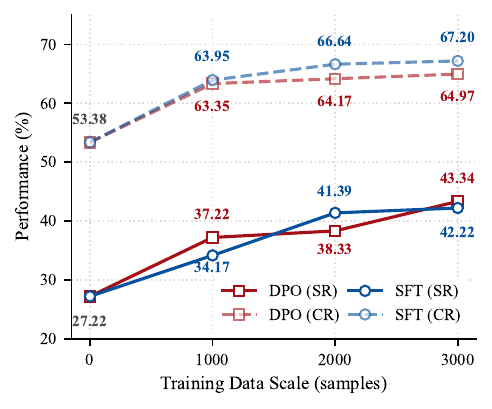}
	\caption{Effect of training data scale on average performance across 120 multi-step tasks. Both CR and SR are reported.}
	\Description{A line chart plots completion rate as dashed lines and success rate as solid lines for SFT and DPO as the training set grows from 0 to 3,000 samples. Completion rates rise from 53.38 percent to 67.20 percent for SFT and 64.97 percent for DPO, while success rates rise from 27.22 percent to 42.22 percent and 43.34 percent, respectively, with DPO overtaking SFT in success rate at 3,000 samples.}
	\label{fig:data_scale}
\end{figure}

We investigate the impact of training data scale on self-evolution performance across 120 multi-step tasks. As illustrated by the learning curves in Figure~\ref{fig:data_scale}, SFT and DPO exhibit distinct scaling behaviors. SFT demonstrates rapid initial convergence with diminishing returns: Success Rate (SR) surges by 7.22\% (from 34.17\% to 41.39\%) when scaling from 1,000 to 2,000 samples, but plateaus at 3,000 (42.22\%). Conversely, while DPO's Completion Rate (CR) gains are marginal, its SR trajectory shows superior scalability, ultimately reaching 43.34\% at 3,000 and outperforming SFT. This divergence suggests that preference-based optimization excels at refining the decision boundary for task success even after parametric learning saturates. Consequently, while 2,000 samples provide a cost-effective equilibrium for SFT, DPO requires larger datasets ($\geq$3,000) to fully manifest its advantages in decision-quality optimization.

\section{Conclusion}
We present \textsc{Clare}, a self-evolving, clarify-before-execute multi-agent framework that resolves underspecified requirements in 3D tool orchestration, together with the 620-task, five-domain 3D-Clarify benchmark. Across static prompting, dynamic agentic methods, and task-interface-matched CLAMBER, CEP, and 3D-GPT adapters, \textsc{Clare} consistently improves execution reliability by clarifying ambiguity, missing information, and mistaken constraints before tool use. Its MR-optimized policy reaches 60.40\% SR on 500 single-step tasks and 43.34\% SR on 120 multi-step tasks, more than doubling the strongest baseline on single-step tasks. These results show that proactive clarification policies can be learned and optimized through simulated multi-turn interaction, offering a practical path toward more reliable human--AI collaboration in specification-heavy 3D workflows.

\paragraph{Limitations.}
The overall execution success of \textsc{Clare} is inherently bounded by the capabilities of the backbone LLM and the integrated 3D toolchains. Additionally, while our simulated environment effectively optimizes clarification policies, fully capturing the diversity and unpredictability of real-world human queries remains an ongoing challenge.

\begin{acks}
	This work was supported in part by the GJYC program of Guangzhou under Grant 2024D01J0081, in part by the ZJ program of Guangdong under Grant 2023QN10X455, and in part by the Fundamental Research Funds for the Central Universities under Grant 2025ZYGXZR-053. Sponsored by CAAI-Lenovo Blue Sky Research Fund.
\end{acks}

\bibliographystyle{ACM-Reference-Format}
\bibliography{refs}

@String{Computer = "{IEEE} Computer" }

@String{Chelsea = "Chelsea" }

@article{wu2025collabllm,
  title={Collabllm: From passive responders to active collaborators},
  author={Wu, Shirley and Galley, Michel and Peng, Baolin and Cheng, Hao and Li, Gavin and Dou, Yao and Cai, Weixin and Zou, James and Leskovec, Jure and Gao, Jianfeng},
  journal={arXiv preprint arXiv:2502.00640},
  year={2025}
}

@article{rafailov2023direct,
  title={Direct preference optimization: Your language model is secretly a reward model},
  author={Rafailov, Rafael and Sharma, Archit and Mitchell, Eric and Manning, Christopher D and Ermon, Stefano and Finn, Chelsea},
  journal={Advances in neural information processing systems},
  volume={36},
  pages={53728--53741},
  year={2023}
}

@inproceedings{lee2024text,
  title={Text-to-3D Shape Generation},
  author={Lee, Hanhung and Savva, Manolis and Chang, Angel X},
  booktitle={Computer Graphics Forum},
  volume={43},
  number={2},
  pages={e15061},
  year={2024},
  organization={Wiley Online Library}
}

@article{jiang2024survey,
  title={A survey on text-to-3d contents generation in the wild},
  author={Jiang, Chenhan},
  journal={arXiv preprint arXiv:2405.09431},
  year={2024}
}

@article{zhang2025advances,
  title={Advances in feed-forward 3d reconstruction and view synthesis: A survey},
  author={Zhang, Jiahui and Li, Yuelei and Chen, Anpei and Xu, Muyu and Liu, Kunhao and Wang, Jianyuan and Long, Xiao-Xiao and Liang, Hanxue and Xu, Zexiang and Su, Hao and others},
  journal={arXiv preprint arXiv:2507.14501},
  year={2025}
}

@inproceedings{yao2023react,
  title = {{ReAct}: Synergizing Reasoning and Acting in Language Models},
  author = {Yao, Shunyu and Zhao, Jeffrey and Yu, Dian and Du, Nan and Shafran, Izhak and Narasimhan, Karthik and Cao, Yuan},
  booktitle = {International Conference on Learning Representations (ICLR) },
  year = {2023},
  html = {https://arxiv.org/abs/2210.03629},
}

@article{shinn2023reflexion,
  title={Reflexion: Language agents with verbal reinforcement learning},
  author={Shinn, Noah and Cassano, Federico and Gopinath, Ashwin and Narasimhan, Karthik and Yao, Shunyu},
  journal={Advances in neural information processing systems},
  volume={36},
  pages={8634--8652},
  year={2023}
}

@article{wei2022chain,
  title={Chain-of-thought prompting elicits reasoning in large language models},
  author={Wei, Jason and Wang, Xuezhi and Schuurmans, Dale and Bosma, Maarten and Xia, Fei and Chi, Ed and Le, Quoc V and Zhou, Denny and others},
  journal={Advances in neural information processing systems},
  volume={35},
  pages={24824--24837},
  year={2022}
}

@misc{sun2025trainingproactivepersonalizedllm,
      title={Training Proactive and Personalized LLM Agents}, 
      author={Weiwei Sun and Xuhui Zhou and Weihua Du and Xingyao Wang and Sean Welleck and Graham Neubig and Maarten Sap and Yiming Yang},
      year={2025},
      eprint={2511.02208},
      archivePrefix={arXiv},
      primaryClass={cs.AI},
      url={https://arxiv.org/abs/2511.02208}, 
}

@misc{qian2025userbenchinteractivegymenvironment,
      title={UserBench: An Interactive Gym Environment for User-Centric Agents}, 
      author={Cheng Qian and Zuxin Liu and Akshara Prabhakar and Zhiwei Liu and Jianguo Zhang and Haolin Chen and Heng Ji and Weiran Yao and Shelby Heinecke and Silvio Savarese and Caiming Xiong and Huan Wang},
      year={2025},
      eprint={2507.22034},
      archivePrefix={arXiv},
      primaryClass={cs.AI},
      url={https://arxiv.org/abs/2507.22034}, 
}

@misc{qian2025userrltraininginteractiveusercentric,
      title={UserRL: Training Interactive User-Centric Agent via Reinforcement Learning}, 
      author={Cheng Qian and Zuxin Liu and Akshara Prabhakar and Jielin Qiu and Zhiwei Liu and Haolin Chen and Shirley Kokane and Heng Ji and Weiran Yao and Shelby Heinecke and Silvio Savarese and Caiming Xiong and Huan Wang},
      year={2025},
      eprint={2509.19736},
      archivePrefix={arXiv},
      primaryClass={cs.AI},
      url={https://arxiv.org/abs/2509.19736}, 
}

@misc{sun20243dgptprocedural3dmodeling,
      title={3D-GPT: Procedural 3D Modeling with Large Language Models}, 
      author={Chunyi Sun and Junlin Han and Weijian Deng and Xinlong Wang and Zishan Qin and Stephen Gould},
      year={2024},
      eprint={2310.12945},
      archivePrefix={arXiv},
      primaryClass={cs.CV},
      url={https://arxiv.org/abs/2310.12945}, 
}

@misc{fang2025comprehensivesurveyselfevolvingai,
      title={A Comprehensive Survey of Self-Evolving AI Agents: A New Paradigm Bridging Foundation Models and Lifelong Agentic Systems}, 
      author={Jinyuan Fang and Yanwen Peng and Xi Zhang and Yingxu Wang and Xinhao Yi and Guibin Zhang and Yi Xu and Bin Wu and Siwei Liu and Zihao Li and Zhaochun Ren and Nikos Aletras and Xi Wang and Han Zhou and Zaiqiao Meng},
      year={2025},
      eprint={2508.07407},
      archivePrefix={arXiv},
      primaryClass={cs.AI},
      url={https://arxiv.org/abs/2508.07407}, 
}

@misc{acikgoz2025macmultiagentframeworkinteractive,
      title={MAC: A Multi-Agent Framework for Interactive User Clarification in Multi-turn Conversations}, 
      author={Emre Can Acikgoz and Jinoh Oh and Joo Hyuk Jeon and Jie Hao and Heng Ji and Dilek Hakkani-Tür and Gokhan Tur and Xiang Li and Chengyuan Ma and Xing Fan},
      year={2025},
      eprint={2512.13154},
      archivePrefix={arXiv},
      primaryClass={cs.AI},
      url={https://arxiv.org/abs/2512.13154}, 
}

@article{acikgoz2025speakrlsynergizingreasoningspeaking,
      title={SpeakRL: Synergizing Reasoning, Speaking, and Acting in Language Models with Reinforcement Learning}, 
      author={Emre Can Acikgoz and Jinoh Oh and Jie Hao and Joo Hyuk Jeon and Heng Ji and Dilek Hakkani-Tür and Gokhan Tur and Xiang Li and Chengyuan Ma and Xing Fan},
      year={2025},
      eprint={2512.13159},
      archivePrefix={arXiv},
      primaryClass={cs.AI},
      url={https://arxiv.org/abs/2512.13159}, 
}

@article{
gao2026a,
title={A Survey of Self-Evolving Agents: What, When, How, and Where to Evolve on the Path to Artificial Super Intelligence},
author={Huan-ang Gao and Jiayi Geng and Wenyue Hua and Mengkang Hu and Xinzhe Juan and Hongzhang Liu and Shilong Liu and Jiahao Qiu and Xuan Qi and Qihan Ren and Yiran Wu and Hongru WANG and Han Xiao and Yuhang Zhou and Shaokun Zhang and Jiayi Zhang and Jinyu Xiang and Yixiong Fang and Qiwen Zhao and Dongrui Liu and Cheng Qian and Zhenhailong Wang and Minda Hu and Huazheng Wang and Qingyun Wu and Heng Ji and Mengdi Wang},
journal={Transactions on Machine Learning Research},
issn={2835-8856},
year={2026},
url={https://openreview.net/forum?id=CTr3bovS5F},
note={Survey Certification}
}

@misc{chen2024idea23dcollaborativelmmagents,
      title={Idea23D: Collaborative LMM Agents Enable 3D Model Generation from Interleaved Multimodal Inputs}, 
      author={Junhao Chen and Xiang Li and Xiaojun Ye and Chao Li and Zhaoxin Fan and Hao Zhao},
      year={2024},
      eprint={2404.04363},
      archivePrefix={arXiv},
      primaryClass={cs.CV},
      url={https://arxiv.org/abs/2404.04363}, 
}

@misc{yang2023llmgrounderopenvocabulary3dvisual,
      title={LLM-Grounder: Open-Vocabulary 3D Visual Grounding with Large Language Model as an Agent}, 
      author={Jianing Yang and Xuweiyi Chen and Shengyi Qian and Nikhil Madaan and Madhavan Iyengar and David F. Fouhey and Joyce Chai},
      year={2023},
      eprint={2309.12311},
      archivePrefix={arXiv},
      primaryClass={cs.CV},
      url={https://arxiv.org/abs/2309.12311}, 
}

@misc{ai2025nli4volvisnaturallanguageinteraction,
      title={NLI4VolVis: Natural Language Interaction for Volume Visualization via LLM Multi-Agents and Editable 3D Gaussian Splatting}, 
      author={Kuangshi Ai and Kaiyuan Tang and Chaoli Wang},
      year={2025},
      eprint={2507.12621},
      archivePrefix={arXiv},
      primaryClass={cs.HC},
      url={https://arxiv.org/abs/2507.12621}, 
}

@misc{zhou2024symboliclearningenablesselfevolving,
      title={Symbolic Learning Enables Self-Evolving Agents}, 
      author={Wangchunshu Zhou and Yixin Ou and Shengwei Ding and Long Li and Jialong Wu and Tiannan Wang and Jiamin Chen and Shuai Wang and Xiaohua Xu and Ningyu Zhang and Huajun Chen and Yuchen Eleanor Jiang},
      year={2024},
      eprint={2406.18532},
      archivePrefix={arXiv},
      primaryClass={cs.CL},
      url={https://arxiv.org/abs/2406.18532}, 
}

@misc{yuan2025agentrtraininglanguagemodel,
      title={Agent-R: Training Language Model Agents to Reflect via Iterative Self-Training}, 
      author={Siyu Yuan and Zehui Chen and Zhiheng Xi and Junjie Ye and Zhengyin Du and Jiecao Chen},
      year={2025},
      eprint={2501.11425},
      archivePrefix={arXiv},
      primaryClass={cs.AI},
      url={https://arxiv.org/abs/2501.11425}, 
}

@misc{zhang2024clamberbenchmarkidentifyingclarifying,
      title={CLAMBER: A Benchmark of Identifying and Clarifying Ambiguous Information Needs in Large Language Models}, 
      author={Tong Zhang and Peixin Qin and Yang Deng and Chen Huang and Wenqiang Lei and Junhong Liu and Dingnan Jin and Hongru Liang and Tat-Seng Chua},
      year={2024},
      eprint={2405.12063},
      archivePrefix={arXiv},
      primaryClass={cs.CL},
      url={https://arxiv.org/abs/2405.12063}, 
}

@misc{chen2024styleimprovingdomaintransferability,
      title={STYLE: Improving Domain Transferability of Asking Clarification Questions in Large Language Model Powered Conversational Agents}, 
      author={Yue Chen and Chen Huang and Yang Deng and Wenqiang Lei and Dingnan Jin and Jia Liu and Tat-Seng Chua},
      year={2024},
      eprint={2405.12059},
      archivePrefix={arXiv},
      primaryClass={cs.CL},
      url={https://arxiv.org/abs/2405.12059}, 
}

@misc{
kudrinskii2026faithful,
title={Faithful Simulation of User{\textendash}Agent{\textendash}Environment Interactions for Scalable {LLM} Agent Evaluation},
author={Aleksei Kudrinskii and Saibo Geng and Luca Beurer-Kellner and Marc Fischer},
year={2026},
url={https://openreview.net/forum?id=dYO3XS9Wsm}
}

@misc{zhang2024askbeforeplanproactivelanguageagents,
      title={Ask-before-Plan: Proactive Language Agents for Real-World Planning}, 
      author={Xuan Zhang and Yang Deng and Zifeng Ren and See-Kiong Ng and Tat-Seng Chua},
      year={2024},
      eprint={2406.12639},
      archivePrefix={arXiv},
      primaryClass={cs.CL},
      url={https://arxiv.org/abs/2406.12639}, 
}

@misc{huang2025teachinglanguagemodelsgather,
      title={Teaching Language Models To Gather Information Proactively}, 
      author={Tenghao Huang and Sihao Chen and Muhao Chen and Jonathan May and Longqi Yang and Mengting Wan and Pei Zhou},
      year={2025},
      eprint={2507.21389},
      archivePrefix={arXiv},
      primaryClass={cs.AI},
      url={https://arxiv.org/abs/2507.21389}, 
}

@article{Zhou2018,
    author    = {Qian-Yi Zhou and Jaesik Park and Vladlen Koltun},
    title     = {{Open3D}: {A} Modern Library for {3D} Data Processing},
    journal   = {arXiv:1801.09847},
    year      = {2018},
}

@article{zhang2025concerto,
  title={Concerto: Joint 2d-3d self-supervised learning emerges spatial representations},
  author={Zhang, Yujia and Wu, Xiaoyang and Lao, Yixing and Wang, Chengyao and Tian, Zhuotao and Wang, Naiyan and Zhao, Hengshuang},
  journal={arXiv preprint arXiv:2510.23607},
  year={2025}
}

@inproceedings{wang2025vggt,
  title={Vggt: Visual geometry grounded transformer},
  author={Wang, Jianyuan and Chen, Minghao and Karaev, Nikita and Vedaldi, Andrea and Rupprecht, Christian and Novotny, David},
  booktitle={Proceedings of the Computer Vision and Pattern Recognition Conference},
  pages={5294--5306},
  year={2025}
}

@misc{lai2025hunyuan3d25highfidelity3d,
      title={Hunyuan3D 2.5: Towards High-Fidelity 3D Assets Generation with Ultimate Details}, 
      author={Tencent Hunyuan3D Team},
      year={2025},
      eprint={2506.16504},
      archivePrefix={arXiv},
      primaryClass={cs.CV},
      url={https://arxiv.org/abs/2506.16504}, 
}

@misc{hunyuan3d22025tencent,
    title={Hunyuan3D 2.0: Scaling Diffusion Models for High Resolution Textured 3D Assets Generation},
    author={Tencent Hunyuan3D Team},
    year={2025},
    eprint={2501.12202},
    archivePrefix={arXiv},
    primaryClass={cs.CV}
}

@misc{yang2024hunyuan3d,
    title={Hunyuan3D 1.0: A Unified Framework for Text-to-3D and Image-to-3D Generation},
    author={Tencent Hunyuan3D Team},
    year={2024},
    eprint={2411.02293},
    archivePrefix={arXiv},
    primaryClass={cs.CV}
}

@article{xiang2024structured,
    title   = {Structured 3D Latents for Scalable and Versatile 3D Generation},
    author  = {Xiang, Jianfeng and Lv, Zelong and Xu, Sicheng and Deng, Yu and Wang, Ruicheng and Zhang, Bowen and Chen, Dong and Tong, Xin and Yang, Jiaolong},
    journal = {arXiv preprint arXiv:2412.01506},
    year    = {2024}
}

@inproceedings{yang2025fast3r,
  title={Fast3r: Towards 3d reconstruction of 1000+ images in one forward pass},
  author={Yang, Jianing and Sax, Alexander and Liang, Kevin J and Henaff, Mikael and Tang, Hao and Cao, Ang and Chai, Joyce and Meier, Franziska and Feiszli, Matt},
  booktitle={Proceedings of the Computer Vision and Pattern Recognition Conference},
  pages={21924--21935},
  year={2025}
}

@article{ma2025progressive,
  title={Progressive Rendering Distillation: Adapting Stable Diffusion for Instant Text-to-Mesh Generation without 3D Data},
  author={Ma, Zhiyuan and Liang, Xinyue and Wu, Rongyuan and Zhu, Xiangyu and Lei, Zhen and Zhang, Lei},
  booktitle={Proceedings of the IEEE/CVF conference on computer vision and pattern recognition},
  year={2025}
}

@misc{worldgen2025ziyangxie,
  author = {Ziyang Xie},
  title = {WorldGen: Generate Any 3D Scene in Seconds},
  year = {2025},
  publisher = {GitHub},
  journal = {GitHub repository},
  howpublished = {\url{https://github.com/ZiYang-xie/WorldGen}},
}

@inproceedings{zheng2024llamafactory,
  title={LlamaFactory: Unified Efficient Fine-Tuning of 100+ Language Models},
  author={Yaowei Zheng and Richong Zhang and Junhao Zhang and Yanhan Ye and Zheyan Luo and Zhangchi Feng and Yongqiang Ma},
  booktitle={Proceedings of the 62nd Annual Meeting of the Association for Computational Linguistics (Volume 3: System Demonstrations)},
  address={Bangkok, Thailand},
  publisher={Association for Computational Linguistics},
  year={2024},
  url={http://arxiv.org/abs/2403.13372}
}

@InProceedings{Jensen_2014_CVPR,
author = {Jensen, Rasmus and Dahl, Anders and Vogiatzis, George and Tola, Engin and Aanaes, Henrik},
title = {Large Scale Multi-view Stereopsis Evaluation},
booktitle = {Proceedings of the IEEE Conference on Computer Vision and Pattern Recognition (CVPR)},
month = {June},
year = {2014}
}

@article{barron2022mipnerf360,
    title={Mip-NeRF 360: Unbounded Anti-Aliased Neural Radiance Fields},
    author={Jonathan T. Barron and Ben Mildenhall and 
            Dor Verbin and Pratul P. Srinivasan and Peter Hedman},
    journal={CVPR},
    year={2022}
}

@article{Knapitsch2017,
    author    = {Arno Knapitsch and Jaesik Park and Qian-Yi Zhou and Vladlen Koltun},
    title     = {Tanks and Temples: Benchmarking Large-Scale Scene Reconstruction},
    journal   = {ACM Transactions on Graphics},
    volume    = {36},
    number    = {4},
    year      = {2017},
}

@inproceedings{dai2017scannet,
    title={ScanNet: Richly-annotated 3D Reconstructions of Indoor Scenes},
    author={Dai, Angela and Chang, Angel X. and Savva, Manolis and Halber, Maciej and Funkhouser, Thomas and Nie{\ss}ner, Matthias},
    booktitle = {Proc. Computer Vision and Pattern Recognition (CVPR), IEEE},
    year = {2017}
}

\clearpage
% Merged appendix for the CLARE arXiv preprint.

\appendix
\section{Method Details}
\label{app:method-details}
\subsection{Architecture}
\label{app:architecture}
To effectively manage the uncertainty of human intent, CLARE employs a multi-agent progressive alignment loop that decouples the 3D generation pipeline into four specialized cognitive roles. This collaborative architecture ensures that requirements are fully specified before invoking computationally expensive 3D tools.
\begin{itemize}
\item Proactive Approver (Cognitive Gatekeeper): Acting as the first line of defense, the Approver evaluates the completeness of the user's current instruction against the multi-turn conversation history. Instead of defaulting to blind execution, it computes a binary decision state to detect ambiguity, missing information, or mistaken details. The Approver outputs its decision in a strict XML format (<judgement> and <reason>), effectively freezing the execution pipeline and categorizing the rejection (e.g., unclear objectives, invalid file paths) until the user's intent becomes fully actionable.
\item Contextual Summarizer (Structured Intent Mining): Once a request is approved, the Summarizer acts as the system's dynamic memory by aggregating fragmented multi-turn dialogue into a normalized task manifest. It performs essential state tracking and co-reference resolution. The output is strictly formatted into XML fields (<task\_type>, <task\_summary>, <file\_path>, and <extra\_info>), which distills noisy conversational context into explicit execution constraints and provides a noise-free specification for the downstream code synthesis.
\item 3D-Aware Programmer (Adaptive Tool Orchestration): The Programmer is responsible for translating the structured manifest into executable Python code. Unlike generic code generators, it incorporates a built-in Model Selection Knowledge Base to understand the capability boundaries and input/output formats of different 3D tools. It dynamically maps user-defined constraints (such as speed and quality) to specific models $\mathcal{T}_{3D}$ and execution parameters $\theta_{exec}$. This ensures that the synthesized code is not only syntactically correct but also physically and semantically grounded in the 3D generation context.
\item Interaction Feedback Provider (Closing the Semantic Gap): This module serves as the interface for conversational alignment, translating internal system states into human-readable, actionable feedback. It operates in two adaptive modes. In Clarification Mode, it intercepts the Approver's rejection reasoning and translates it into targeted follow-up questions, guiding the user to provide missing parameters without cognitive overload. In Status Reporting Mode, it interprets raw execution logs and potential runtime errors post-execution, providing the user with concise summaries and clear next-step suggestions to iteratively refine their 3D assets.
\end{itemize}
\subsection{Details and Prompts of CLARE Agents}
\label{app:agent-prompts}
\subsubsection{Proactive Approver}
The Approver is the core role responsible for "decision gating." In each round of dialogue, it receives the current user input and the existing dialogue history, combines them into a complete requirement, and then determines whether the requirement meets the "executable" condition. "Executable" means that the requirement must contain a clear task objective (four types of 3D workflows), corresponding input information (such as image file paths or text prompts), and necessary parameters and output format. In addition, for generation/reconstruction classes, a clear output format is required (such as .ply/.glb), for post-processing classes, specific operations are required (such as removing outliers/converting meshes to point clouds), and for point cloud editing classes, the "detail objects" to be edited are required (such as "remove chair/keep table"). When the task is sufficiently specific, Approver must output strict XML: `<judgement> Yes</judgement>`. If the task still contains missing, ambiguous, or erroneous elements, it must output `<judgement> No</judgement>` along with a `<reason>` in the same output, where the reason is selected from predefined rejection categories (Missing information / Task objective unclear / Mistaken details / User wants to clarify or confirm the task details). Furthermore, Approver's output is protocolized to a format that "must contain only these XML tags," ensuring stable and reliable routing. Assistants, as the core of the Agent framework, support the use of trained preference models (e.g., via DPO/SFT), allowing them to directly influence the quality of gating decisions.
\begin{tcolorbox}[title=Proactive Approver, breakable, colback=bluebg, colbacktitle=bluebg, colframe=bluefg, fonttitle=\bfseries\color{black}, width=\linewidth]
\begin{Verbatim}[breaklines=true, breakanywhere=true, fontsize=\footnotesize]
# TASK
You are the Approver in an interactive multimodal 3D Agent system.
Your responsibility is to:
1. Receive the conversation history and the user's current request.
2. Combine all history messages and current request into one complete demand.
3. Carefully read and interpret the user's request with all history information.
4.Judge whether the user's input (combined with all history messages) already provides enough information to generate a concrete task.

   - If YES, you must output in the following XML structure:
    <judgement> Yes </judgement>

   - If NO, you must output in the following XML structure:
    <judgement> No </judgement>
    <reason> [**Choose** one or some of the categories below and **Explain** clearly what is missing or mistaken] </reason>

# Reject Categories:
1. Missing information
   - The user described an input source (image/video/point cloud) but did not provide the actual input file path or output path.
   - The user mention but did not specify value for each parameter.

2. Task objective unclear
   - The user did not specify the intended operation (e.g., generation / post processing / reconstruction / editing) or desired output format, making the task ambiguous.

3. Mistaken details
   - The user provided contradictory details that cannot be executed. (e.g. Mistaken input or output formats / incompatible parameters / conflicting properties)
   - The user provided an input file path that does not match the described task objective. (e.g. image path for point cloud processing tasks)
   
4. User wants to clarify or confirm the task details


# Rules:
A task is considered concrete if it includes:
- A clear objective (can be classified into **one of or the combination of** the following 4 categories):
  * Single-View / Multi-View 3D Reconstruction (Single view from image / Multi view from images or video)
  * Text to 3D Generation (e.g., object or scene)
  * Post Processing (e.g., remove outliers, mesh to pointcloud conversion, etc.)
  * Point Cloud Editing (e.g. remove chair, keep table, etc. must have detail object to edit)
- **A specific input file path** (for image/video/point cloud inputs, or a text prompt for text-based generation).
- Any necessary parameters for each task (e.g., optional parameters for outliers removal, output file's format, etc.)
For 3D Reconstruction and Text to 3D Generation tasks, clairly specify the desired output format (e.g., file format '.ply', '.glb', object format mesh / pointcloud etc.)
For Post Processing tasks, specify the exact operation to be performed on the input data (e.g., "remove outliers" or "mesh to point cloud conversion").
For Point Cloud Editing tasks, clearly identify the *detail objects* to be edited (e.g., "remove the chair", "keep the table").

# Other Rules:
- Always output in strict XML tags as specified.
- Never output anything else except the XML tags.
- Users DO NOT need to specify which category the task belongs to.
- If the user mentions generating and does not provide any other image/video input, it is assumed to be a text to mesh/pcd task without asking again about the task objective.
- If the user does not specify the task objective or necessary parameters for each task, do not approve it.
- If the user does not specify the input file path for image/video/point cloud tasks, do not approve it.
- If the user ask to clarify or modify the task details, response the reason like this to let other module answer: <reason> The user is trying to confirm or clarify the task details. Please answer the user. </reason>
\end{Verbatim} 
\end{tcolorbox}
\subsubsection{Contextual Summarizer}
Once the Approver determines that the current dialogue is sufficiently specific, the system calls the Summarizer to compress the multi-turn dialogue into an "executable, structured task description." This summary is output in strict XML format and contains four fixed fields: `\texttt{task\_type}` (supporting combinations such as reconstruction/generation/post-processing/point cloud editing), `\texttt{task\_summary}` (breaking down the requirements into concise steps and describing the output format of each step), `\texttt{file\_path}` (a set of user-provided input paths), and `\texttt{extra\_info}` (execution-related parameters and preference information, such as quality/speed/style/output format). This structured summary is then packaged as input and passed to the Programmer to generate the final executable code.
\begin{Verbatim}[breaklines=true, breakanywhere=true, fontsize=\footnotesize]
<task_type> Multi-View 3D Reconstruction </task_type>
<task_summary> Perform multi-view 3D reconstruction using ./data/source/multiview/scene/Auditorium as the input source based on constraints </task_summary>
<file_path> ./data/source/multiview/scene/Auditorium </file_path>
<extra_info> Constraints: [speed: Fast speed]; Output Path: ./outputs/Auditorium.ply </extra_info>
\end{Verbatim}

\begin{tcolorbox}[title=Contextual Summarizer, breakable, colback=bluebg, colbacktitle=bluebg, colframe=bluefg, fonttitle=\bfseries\color{black}, width=\linewidth]
\begin{Verbatim}[breaklines=true, breakanywhere=true, fontsize=\footnotesize]
# TASK
You are the Summarizer in an interactive multimodal 3D Agent system. 
Your responsibility is to:
1. Combine **all** history messages and current request into complete demand even if some tasks have been completed in previous steps. 
2. Summarize all the demand using the following XML structure:
    <task_type> [Summarized and structured task name, e.g. 3D Reconstruction (Multi-View), Point Cloud Editing, Post Processing (mesh to point cloud) etc.] </task_type>
    <task_summary> [Concise summary of the task after decomposition into steps] </task_summary>
    <file_path> [user-provided file path(s) | None if not provided] </file_path>
    <extra_info> [Any useful additional context or None] </extra_info>

# Rules:
task_type tag includes:
- A clear objective (only includes **one of or the combination of** the following 4 types):
  * Single-View / Multi-View 3D Reconstruction (Single view from image / Multi view from images or video)
  * Text to 3D Generation (e.g., object or scene)
  * Post Processing (including outliers removal, mesh to point cloud conversion)
  * Point Cloud Editing (e.g. remove chair, keep table, etc.)
task_summary tag includes:
- A brief steps after decomposition of the task, avoiding unnecessary details but include the output type of each step.
file tag includes:
- Specific input details (e.g., text prompt for generation tasks, image or video file path for reconstruction tasks, etc.)
extra_info tag includes:
- Any necessary parameters (e.g., optional parameters for outliers removal, output file's format, etc.)
- Preserve ALL adjectives related to quality, speed, style, or format (e.g., "high quality", "fast", "low poly") in <extra_info> and <task_summary>.

# Other Rules:
- Always output in strict XML tags as specified.
- Never output anything else except the XML tags.
- For multiple tasks, separate them with commas.
- Prioritize continuing from previous results, that is, avoid repeating completed steps.
- If the previous results cannot be utilized, summarize from scratch.
- Always include **all** tasks in your output even if some tasks have been completed in previous steps.

# Examples:
<example1>
History: None.
User: Perform multi-view 3D reconstruction using ./data/source/multiview/scene/Auditorium as the input source. I need it processed with fast speed. Save the output to ./outputs/Auditorium.ply.
Assistant (you): <task_type> Multi-View 3D Reconstruction </task_type>
            <task_summary> Perform multi-view 3D reconstruction using ./data/source/multiview/scene/Auditorium as the input source based on constraints </task_summary>
            <file_path> ./data/source/multiview/scene/Auditorium </file_path>
            <extra_info> Constraints: [speed: Fast speed]; Output Path: ./outputs/Auditorium.ply </extra_info>
</example1>

<example2>
History: None.  
User: Perform single-view 3D reconstruction using ./data/source/singleview/object/scan11.jpg as input. Expect medium speed processing. Save the output as a ply file. Save the output to ./outputs/scan11.ply.\nThen, please process the mesh at the output path of the previous step to create a point cloud. Use a sampling rate of 25000 points and write the file to ./outputs/example_mesh_from_ply_25000pts.ply.\nFinally, clean the point cloud file the output path of the previous step. I want to remove outliers with nb_neighbors set to 30 and standard deviation ratio set to 0.9. Output result to ./outputs/clean_nb30_std0.9.ply.
Assistant (you): <task_type> Single-View 3D Reconstruction, Post Processing, Post Processing </task_type>
         <task_summary> Perform single-view 3D reconstruction using ./data/source/singleview/object/scan11.jpg as input. Expect medium speed processing. Save the output as a ply file. Save the output to ./outputs/scan11.ply.\nThen, please process the mesh at the output path of the previous step to create a point cloud. Use a sampling rate of 25000 points and write the file to ./outputs/example_mesh_from_ply_25000pts.ply.\nFinally, clean the point cloud file the output path of the previous step. I want to remove outliers with nb_neighbors set to 30 and standard deviation ratio set to 0.9. Output result to ./outputs/clean_nb30_std0.9.ply. </task_summary>
         <file_path> Input 1:./data/source/singleview/object/scan11.jpg; Input 2:./outputs/scan11.ply; Input 3:./outputs/example_mesh_from_ply_25000pts.ply;  </file_path>
         <extra_info> Step 1: { Constraints: [speed: Medium speed, output_format: ply]; Output Path: ./outputs/scan11.ply }; Step 2: { number_of_points: 25000; Output Path: ./outputs/example_mesh_from_ply_25000pts.ply }; Step 3: { nb_neighbors: 30, std_ratio: 0.9, Output Path: ./outputs/clean_nb30_std0.9.ply } </extra_info>
</example2>

<example3>
History: User: I want to make a glass 20-sided die. Make sure it's high quality. The output should be in PLY format. Assistant: The task was successfully completed. A high-quality 20-sided glass die was generated and saved as a PLY file at the path "./tmp/high_quality_20_sided_die.ply". You can now use this file for further processing or 3D printing. If you need any modifications or additional steps, feel free to ask!
User: Now, take the mesh from that file and generate a point cloud with 15000 samples. Save it to ./outputs/example_mesh_from_ply_15000pts.ply.
Assistant (you): <task_type> Text to 3D Generation, Post Processing </task_type>
         <task_summary> Generate a 3D model of a glass 20-sided die. Ensure the result is high quality. The generation speed should be medium speed. Consider that this is a high fidelity mesh. Save the output to ./outputs/a_glass_20-sided_die.ply.\nThen, take the mesh at the output path of the previous step and generate a point cloud with 15000 samples. Store the output at ./outputs/example_mesh_from_ply_15000pts.ply. </task_summary>
         <file_path> Input 1:None; Input 2:./outputs/a_glass_20-sided_die.ply;  </file_path>
         <extra_info> Step 1: { Constraints: [quality: High quality, speed: Medium speed, features: High fidelity mesh]; Output Path: ./outputs/a_glass_20-sided_die.ply }; Step 2: { number_of_points: 15000; Output Path: ./outputs/example_mesh_from_ply_15000pts.ply } </extra_info>
</example3>
\end{Verbatim} 
\end{tcolorbox}

\subsubsection{3D-Aware Programmer}
After receiving the structured task summary, the Programmer first parses the task type and retrieves candidate model information matching the current task type from the built-in Model Selection Knowledge Base. It then injects the output format and capability features of these models as priors for model selection into the prompts, constraining the LLM to follow correct tool selection and parameter constraints when generating code. Subsequently, it generates a directly executable Python code call sequence according to preset template rules, and performs post-processing with regularization to correct key call forms, thereby improving the executability of the generated code and its consistency with downstream 3D tool interfaces.
\begin{tcolorbox}[title=3D-Aware Programmer, breakable, colback=bluebg, colbacktitle=bluebg, colframe=bluefg, fonttitle=\bfseries\color{black}, width=\linewidth]
\begin{Verbatim}[breaklines=true, breakanywhere=true, fontsize=\footnotesize]
# TASK
You are the Programmer in an interactive multimodal 3D Agent system. 
Your responsibility is to:
1. Receive a concrete task (already validated by Approver).
2. Translate the task into correct Python code that calls the Executor class functions.
3. Ensure the generated code is executable without extra modification.

# Rules:
- Always match the correct function to the user's task type (see function descriptions below).
- Always fill in all required arguments.
- Optional arguments (parameters, output_path) should be explicitly set only if specified in the task; otherwise, use default values shown below.
- Honor any output_path or model preference provided in the task (including extra_info hints).
- If no explicit output_path is provided, derive one by slugifying the prompt (lowercase, replace non-alphanumeric characters with underscores, collapse repeats), and use "./tmp/{{slug}}.{{ext}}" where the extension comes from the requested format (default glb if unspecified). Ensure `output_path` is ALWAYS present in the generated code (either from user input or your derived value).
- Never call executor methods without the self.executor prefix (self.method(...) is forbidden).
- Only output the final Python code. Do not explain, do not add comments, do not wrap with markdown.
- Always enforce the user's requested output format: pick a model that natively supports that format.
- If the user's requested format is NOT supported by the selected model (check the Model Selection Guide), you MUST override it to the model's default format and note this change.
- Always call executor functions via self.executor.<function>(...).

# Most Relevant Executor Class Functions:
{executor_description}

# Model Selection Guide:
{SOME_OF_TASK_MODEL_PRIOR}
\end{Verbatim}
\end{tcolorbox}

\begin{tcolorbox}[title=Executor Description, breakable, colback=bluebg, colbacktitle=bluebg, colframe=bluefg, fonttitle=\bfseries\color{black}, width=\linewidth]
\begin{Verbatim}[breaklines=true, breakanywhere=true, fontsize=\footnotesize]
"text_to_3d_generation": 
    """self.executor.text_to_3d_generation(prompt: str, model: str, output_path: str)
    - Usage: Generate a 3D object or scene from a text description.
    - Args:
    	prompt (str): Text prompt describing the target object or scene, e.g. "a futuristic glass chair".
    	model (str, optional): Model to use. Options: "triplaneturbo", "trellis" or "worldgen". Default is "triplaneturbo".
    	output_path (str, optional): Output file path. Defaults to "./tmp/sample_text.ply". "triplaneturbo" will force .obj output; "trellis" supports .ply/.obj/.glb.""",
"single_view_reconstruction":
    """self.executor.single_view_reconstruction(image_path: str, output_path: str, model: str, conf_thres: float)
    - Usage: Reconstruct a 3D object or scene from a single image.
    - Args:
    	image_path (str): Path to the input image.
	    output_path (str, optional): Path for saving the output file. Default is "./tmp/pcd.ply".
        model (str, optional): Model to use. Options: "vggt", "hunyuan", or "trellis". Default is "vggt".
        conf_thres (float, optional): Confidence threshold for filtering low-quality points. Default is 20. Not used for "hunyuan" or "trellis".
        Note: "hunyuan" model (Hunyuan3D-2) only supports image-to-3D generation and outputs 3D mesh without texture. Supports .glb and .obj output formats.""",
"multi_view_reconstruction":
    """self.executor.multi_view_reconstruction(video_path: str, output_path: str, model: str, conf_thres: float)
    - Usage: Reconstruct a 3D scene from multiple view images or a video sequence.
    - Args:
        video_path (str): Path to the input video or multi-view images' folder.
        output_path (str, optional): Path for saving the output file. Default is "./tmp/pcd.ply".
        model (str, optional): Model to use. Options: "vggt", "fast3r", "trellis", or "hunyuan". Default is "vggt".
        conf_thres (float, optional): Confidence threshold for filtering low-quality points. Default is 20. Not used for "trellis" or "hunyuan".""",

"post_processing":
    """self.executor.post_processing(input_path: str, process_type: str, output_path: str, parameters: dict)
    - Usage: Perform post-processing on a point cloud or mesh, such as outlier removal, or conversions.
    - Args:
        input_path (str, optional): Path to the input point cloud or mesh.Default is None.
        process_type (str): Operation type. Options:
            - "outliers_removal": Remove statistical outliers.
            Parameters: {"nb_neighbors": int (default: 20), "std_ratio": float (default: 1.0)}
            - "mesh_to_pointcloud": Convert mesh to point cloud.
            Parameters: {"number_of_points": int (default: 10000)}
        output_path (str, optional): Output file path. Default is "./tmp/pcd_processed.ply".
        parameters (dict, optional): Additional parameters for processing operations. Default is None.""",
            
"pcd_edit":
    """self.executor.pcd_edit(pointcloud_path: str, edit_parts: List[str], output_path: str, edit_type: str)
    - Edit a point cloud by removing, keeping, or segmenting semantic parts.
    - Args:
        pointcloud_path (str): Input point cloud file (.ply).
        edit_parts (List[str]): List of part names to edit (e.g. ["chair", "table"]). If edit_type is "segment", this parameter should be ["all"].
        output_path (str, optional): Output file path. Default is "./tmp/pcd_edited.ply".
        edit_type (str, optional): Editing mode. Options: "delete", "keep", or "segment". Default is "segment".""",
\end{Verbatim}
\end{tcolorbox}

\begin{tcolorbox}[title=Model Selection Knowledge Base, breakable, colback=bluebg, colbacktitle=bluebg, colframe=bluefg, fonttitle=\bfseries\color{black}, width=\linewidth]
\begin{Verbatim}[breaklines=true, breakanywhere=true, fontsize=\footnotesize]
TASK_MODEL_PRIOR = {
    "Text to 3D Generation": [
        {"name": "trellis", "output_format": ["ply", "obj", "glb"], "features": [
            "High quality",
            "Medium speed",
            "Usage: Single object",
            "Additional: High fidelity"
        ]},
        {"name": "triplaneturbo", "output_format": ["obj"], "features": [
            "Low quality",
            "Fast speed",
            "Usage: Single object",
            "Additional: Lightweight, Simplified model"
        ]},
        {"name": "worldgen", "output_format": ["ply", "obj", "glb", "stl"], "features": [
            "Medium quality",
            "Slow speed",
            "Usage: Scene",
            "Additional: Detailed geometry",
        ]},
    ],
    "Single-View 3D Reconstruction": [
        {"name": "vggt", "output_format": ["ply", "glb"], "features": [
            "Medium quality",
            "Fast speed",
            "Usage: Scene or object",
            "Additional: Dense point cloud"
        ]},
        {"name": "hunyuan", "output_format": ["glb", "obj"], "features": [
            "Ultra-High quality",
            "Slow speed",
            "Usage: Single object",
            "Additional: Mesh with detailed geometry"
        ]},
        {"name": "trellis", "output_format": ["ply", "obj", "glb"], "features": [
            "High quality",
            "Medium speed",
            "Usage: Single object",
            "Additional: High-fidelity mesh"
        ]},
    ],
    "Multi-View 3D Reconstruction": [
        {"name": "vggt", "output_format": ["ply", "glb"], "features": [
            "Ultra-High quality",
            "Fast speed",
            "Usage: Scene or object",
            "Additional: Dense point cloud, High resource usage"
        ]},
        {"name": "fast3r", "output_format": ["ply"], "features": [
            "Medium quality",
            "Extreme Fast speed",
            "Usage: Scene or object",
            "Additional: Efficient,Lightweight Point cloud, Low resource usage"
        ]},
        {"name": "trellis", "output_format": ["ply", "obj", "glb"], "features": [
            "High quality",
            "Medium speed",
            "Usage: Single object",
            "Additional: High-fidelity mesh"
        ]},
    ],
    "Post Processing":[
        {"name": "outlier_removal", "output_format": ["ply"], "features": [
            "Remove outliers from point clouds",
        ]},
        {"name": "mesh_to_pointcloud", "output_format": ["ply"], "features": [
            "Convert mesh to point cloud",
        ]},
    ],
    "Point Cloud Editing": [
        {"name": "pcd_edit", "output_format": ["ply"], "features": [
            "Keep indoor objects",
            "Delete indoor objects",
            "Segment indoor objects",
        ]},
    ],
}
\end{Verbatim} 
\end{tcolorbox}

\subsubsection{Interaction Feedback Provider}
The Feedback Provider is responsible for translating internal states into user-facing next-step instructions. When subsequent code execution successfully triggers the system to the `success` state, it provides a brief confirmation and summary to the user based on the input details of the current task and the output location of the generated/executed results, and offers possible next-step suggestions if necessary (e.g., whether further processing or editing is required). Conversely, when the system is in the `failed` state triggered by code execution failure, it identifies the most likely point of error based on the provided error details and the reason for the failure, and provides actionable corrective directions. This helps the user clarify what information needs to be supplemented or which parameters/constraints need to be adjusted, allowing the conversation to quickly return to a recoverable track and continue converging after a failure.
\begin{tcolorbox}[title=Interaction Feedback Provider, breakable, colback=bluebg, colbacktitle=bluebg, colframe=bluefg, fonttitle=\bfseries\color{black}, width=\linewidth]
\begin{Verbatim}[breaklines=true, breakanywhere=true, fontsize=\footnotesize]
# TASK
You are the Feedback Provider in an interactive multimodal 3D Agent system.
Your responsibility is to:
1. Receive the conversation history and the user's original request.
- `history`: A list of alternating user and assistant messages.
2. Receive the latest information about the current task state - it could be a structured message that includes:
- `status`: "success" (by programmer) / "failed" (by programmer) / "rejected" (by approver) / "accepted" (by approver)
- `details`: Detailed information such as task summary, executable code, execution log, error reason, or evaluation comment.
3. Analyze this information carefully and provide **clear, concise, and constructive feedback** to the user to help them understand the outcome and next step. You must respond with:
<feedback>
Your summarized feedback to **the user**.
- If success: summarize the result, mention the **input details (input file path and parameters)** and where to find the output, and possibly suggest next steps or improvements.
- If failed: identify the likely reason, and suggest possible fixes or what the user can clarify next.
- If rejected: explain what the approver found unsatisfactory and what should be revised. If the user wants to clarify or confirm the task details, mention the details.
- If accepted: confirm that the task is approved and ready for execution. Also repeat the complete details for clarity.
</feedback>

# RULES
1. Always communicate in a **helpful, professional, and encouraging tone**.
2. If the output is successful, provide brief validation and next-step suggestions.
3. If it failed or was rejected, summarize the issue and propose or clarify what the user wants **specific, actionable improvements**.
4. Do not include system or internal reasoning in your output - only user-facing feedback.
5. Avoid repeating raw logs; extract meaning from them for the user.
6. Keep the feedback concise but informative (ideally 1 - 3 sentences). 
\end{Verbatim}
\end{tcolorbox}

\subsection{Tools}
\label{app:tools}
We provide a comprehensive suite of 3D processing tools to support diverse agent workflows. Each tool is integrated into the system with specific input/output constraints and functional roles as detailed below:
\paragraph{Trellis}~\cite{xiang2024structured}
\begin{itemize}
\item \textbf{Core Capability:} High-quality 3D generation and reconstruction featuring high-fidelity mesh output.
\item \textbf{System Role:} Serving as the primary model for Text to 3D Generation and 3D Reconstruction.
\item \textbf{I/O:} Supports \texttt{Text/Image} as input and exports to \texttt{.ply}, \texttt{.obj}, or \texttt{.glb} formats.
\end{itemize}\paragraph{TriplaneTurbo}~\cite{ma2025progressive}
\begin{itemize}
\item \textbf{Core Capability:} Lightweight and fast-speed text-to-3D generation.
\item \textbf{System Role:} As a lightweight Text to 3D Generation model.
\item \textbf{I/O:} Processes \texttt{Text} prompts into \texttt{.obj} mesh files.
\end{itemize}\paragraph{WorldGen}~\cite{worldgen2025ziyangxie}
\begin{itemize}
\item \textbf{Core Capability:} Specialized in scene-level 3D generation with detailed geometry.
\item \textbf{System Role:} Handles complex environment generation tasks.
\item \textbf{I/O:} Supports \texttt{Text} input and outputs \texttt{.ply}, \texttt{.obj}, \texttt{.glb}, or \texttt{.stl}.
\end{itemize}\paragraph{VGGT}~\cite{wang2025vggt}
\begin{itemize}
\item \textbf{Core Capability:} Dense 3D reconstruction optimized for both objects and scenes.
\item \textbf{System Role:} Provides fast-speed 3D Reconstruction from single or multiple view inputs, typically producing dense point clouds.
\item \textbf{I/O:} Accepts \texttt{Images} and generates \texttt{.ply} or \texttt{.glb} files.
\end{itemize}\paragraph{Hunyuan-3D}~\cite{lai2025hunyuan3d25highfidelity3d,hunyuan3d22025tencent,yang2024hunyuan3d}
\begin{itemize}
\item \textbf{Core Capability:} Ultra-high quality single-view reconstruction with detailed mesh geometry.
\item \textbf{System Role:} Used for high-precision object digitizing tasks where processing speed is less critical than surface accuracy.
\item \textbf{I/O:} Converts a single \texttt{Image} into \texttt{.glb} or \texttt{.obj} formats.
\end{itemize}\paragraph{Fast3R}~\cite{yang2025fast3r}
\begin{itemize}
\item \textbf{Core Capability:} Extreme-speed multi-view reconstruction using lightweight point cloud representations.
\item \textbf{System Role:} Designed for real-time or low-resource scenarios requiring efficient scene reconstruction.
\item \textbf{I/O:} Processes \texttt{Multi-view Images} into \texttt{.ply} point clouds.
\end{itemize}
\paragraph{Concerto}~\cite{zhang2025concerto}
\begin{itemize}
\item \textbf{Core Capability:} Enables semantic editing of indoor point clouds, including object segmentation, deletion, or retention.
\item \textbf{System Role:} Used for point cloud editing.
\item \textbf{I/O:} Processes \texttt{.ply} point clouds into \texttt{.ply} point clouds.
\end{itemize}
\paragraph{Open3D}~\cite{Zhou2018}
\begin{itemize}
\item \textbf{Outlier Removal:} Statistically cleans noise from raw point clouds to improve downstream quality; supports \texttt{.ply} I/O.
\item \textbf{Mesh2PCD:} Provides geometric conversion by sampling discrete point clouds from mesh surfaces (\texttt{.obj/.glb} to \texttt{.ply}).
\end{itemize}

\subsection{Training Details}
\label{app:training-details}
\subsubsection{Interactive Forward Sampling}

To estimate the long-term utility of a clarification strategy, \textsc{Clare} performs forward lookahead via simulation. Given a context $x = (q_j, \mathcal{H}_{<j})$, the agent samples $N$ candidate responses $\{m^{(1)}, \dots, m^{(N)}\}$ from its current policy $\pi_\theta$. For each candidate, we roll out a full interaction trajectory $\tau^{(i)}$ by coupling the agent with a \textit{Simulated User} $\mathcal{U}$. This process reveals downstream consequences, helping the agent separate vital clarifications from redundant, costly queries. Below, we detail the specific implementation of this simulation loop.

\paragraph{Candidate Generation and Deduplication} 
At each step, the \textbf{Proactive Approver} evaluates the completeness of the user instruction. To encourage exploration of diverse clarification strategies, we sample $N=8$ candidates from the policy using a high temperature ($T=1.0$). Each candidate implies a binary decision state $d \in \{\text{EXECUTE}, \text{CLARIFY}\}$, alongside a corresponding reasoning trace. To optimize computational efficiency during the rollout phase, we apply a deduplication mechanism based on a content hash of the parsed decision and reasoning. Only unique candidates are retained for full trajectory rollouts, avoiding redundant simulations of identical strategies.

\paragraph{Trajectory Rollout via Multi-Role Simulation}
For each unique candidate, we roll out an interaction trajectory $\tau$ that can span up to a maximum limit of $T_{max}=8$ turns. The state transitions depend on the Approver's decision state $d$:
\begin{itemize}
    \item \textbf{Clarification Branch ($d = \text{CLARIFY}$):} The \textbf{Interaction Feedback Provider} translates the Approver's reasoning into a targeted clarification question. The \textbf{Simulated User} $\mathcal{U}$-which holds the ground-truth goal $\mathcal{G}^* = \{p_{in}, f, c, \theta\}$ and operates under a strict information asymmetry setting-provides a response based on its current behavioral stage. This refined input is appended to the dialogue history, and the agent is queried again in the next turn.
    \item \textbf{Execution Branch ($d = \text{EXECUTE}$):} The clarification loop breaks. The \textbf{Contextual Summarizer} aggregates the multi-turn dialogue history into a normalized task manifest $\mathcal{M}$. This manifest is passed to the \textbf{3D-Aware Programmer}, which maps the constraints to specific 3D tools $\mathcal{T}_{3D}$ and parameters $\theta_{exec}$ to synthesize executable Python code.
\end{itemize}

\paragraph{Multi-turn Reward Evaluation and Back-propagation}
Upon trajectory termination (either via code execution or reaching $T_{max}$), an automated verifier evaluates the final output. Rather than a simple binary outcome, we compute a continuous \textbf{Multi-turn Reward (MR)} that balances task fidelity against interaction efficiency:
$$MR(\tau) = \mathcal{V}(\text{Exec}(\mathcal{T}_{3D}, \theta_{exec}), \mathcal{G}^*) - \lambda \cdot K$$
where $K$ is the number of interaction turns and $\lambda$ is a penalty coefficient (set to $0.1$) to discourage verbosity. The value function $\mathcal{V} \in [0, 1]$ enforces strict parametric accuracy by checking both module alignment and specific parameter similarities against the latent goal $\mathcal{G}^*$. 

Because intermediate clarification turns do not yield immediate functional 3D outputs, we employ outcome reward back-propagation: the final trajectory reward $MR(\tau)$ is retroactively mapped to \textit{all} step-level instruction-output pairs within that specific trajectory $\tau$. This mechanism ensures that early, proactive questions are correctly credited for enabling successful downstream execution, providing dense preference signals for subsequent Supervised Fine-Tuning (SFT) or Direct Preference Optimization (DPO).

\subsubsection{Self-Evolution Training Details}
To implement the self-evolution policy optimization described in Section [Insert Main Section Reference], we utilize the LLaMA-Factory framework to train the Proactive Approver. Both the Supervised Fine-Tuning (SFT) and Direct Preference Optimization (DPO) stages share a foundational setup, initialized from the \texttt{Qwen3-8B} backbone. 

\paragraph{Shared Hyperparameters}
To achieve memory-efficient training on our hardware setup (two NVIDIA RTX 3090 GPUs), we employ Low-Rank Adaptation (LoRA)  with a rank of $r=16$, targeting all linear modules (\texttt{lora\_target: all}). Both SFT and DPO are trained for 2 epochs using a peak learning rate of $5.0 \times 10^{-6}$ with a cosine learning rate scheduler and a warmup ratio of $0.1$. We use a per-device batch size of 2 with no gradient accumulation, resulting in an effective total batch size of 4 across the two GPUs. All training is conducted using \texttt{bfloat16} mixed precision and a maximum sequence length (cutoff length) of 2048 tokens to accommodate the multi-turn conversational history.

\paragraph{Supervised Fine-Tuning (SFT) Setup}
For the SFT phase, we fine-tune the agent on a curated dataset of approximately 3,000 high-quality trajectories (filtered by Multi-turn Reward). To optimize GPU memory and computation, this stage utilizes DeepSpeed ZeRO Stage 2 optimization. The objective here is to clone the behavior of the top-ranked clarification trajectories via standard maximum likelihood estimation.

\paragraph{Direct Preference Optimization (DPO) Setup}
For the DPO phase, the dataset consists of approximately 3,000 pairwise comparisons constructed from the simulated rollouts, where "chosen" and "rejected" trajectories are distinguished based on their composite Multi-turn Reward. We employ the standard Sigmoid preference loss. The KL divergence penalty coefficient is set to $\beta=0.1$ to prevent the model from deviating too far from the reference policy. Because DPO requires maintaining both the active policy model and a frozen reference model in memory simultaneously, we scale our memory optimization by utilizing DeepSpeed ZeRO Stage 3 for this phase.

\section{3D Clarify Benchmark Details}
\label{app:benchmark-details}

\subsection{System Prompts for Simulated User}
\label{app:simulated-user}
\begin{tcolorbox}[title=Simulated User, breakable, colback=bluebg, colbacktitle=bluebg, colframe=bluefg, fonttitle=\bfseries\color{black}, width=\linewidth]
\begin{Verbatim}[breaklines=true, breakanywhere=true, fontsize=\footnotesize]
# TASK
You are a simulated user participating in a user study for a multimodal 3D agent. 
Your goal is to test the agent's ability to handle **imperfect, vague, and fragmented** human instructions.
IF you obtain **all** the necessary information from the agent to complete the task, you can end the interaction by saying "Quit" (even if in early stages).

# BEHAVIORAL STAGES (Follow this logic strictly)
1. **Early Stage (Turns 1-2)**: 
   - Be vague and lazy. 
   - If the task requires a file path and a format, ONLY provide the file path initially.
   - Use layman terms.
   - Do NOT reveal the full detailed requirement yet.

2. **Middle Stage (Turns 3-4)**:
   - Provide information only when explicitly asked. If the Agent asks for multiple items (e.g., parameters AND path), answer ALL of them in this turn.
   - You can verify information, but do not volunteer extra details that were NOT asked for.
   - Occasionally simulate a small misunderstanding or clarify a detail.

3. **Final Stage (Turns 5+)**:
   - Now you can be precise to ensure the task is completed.
   - Only say "Quit" when the Agent has confirmed ALL parts of your detailed requirement.

4. **Always**: 
   - Check carefully whether the Agent's feedback confirms it has accepted your detailed requirement (input file path, output path, format, and any parameters). If not, continue the interaction. 
   - Check one by one whether the paths and parameters appearing in the feedback are completely consistent with the detailed requirement.
# CRITICAL RULES
1. **Output ONLY the user response** inside <user> tags.
2. **NEVER** copy-paste the full <detailed_requirement> in one turn. This is a failure.
3. **Goal**: Drag the conversation to at least 4-6 turns if possible by withholding information naturally.
4. **Consistency**: Eventually, you MUST provide all the info in <detailed_requirement>, but do it slowly.
5. End with <user> Quit </user> ONLY when the Agent has explicitly confirmed it has accepted ALL parts of your <detailed_requirement>.
6. Keep responses short and natural, like a real human.
7. Do not add any information that was not provided to you in the detailed task during the conversation. If asked, say use default settings.
8. Any information you provide must appear in the Detailed Requirement.
9. Do not create additional requirements beyond the Detailed Requirement.
10. Do not use keywords that involve other tasks, such as 'clean it up' when editing point clouds, as this can confuse with outlier removal.
11. Avoid repeating yourself.
12. Directly provide necessary information instead of asking for confirm if the feedback does not provide enough information.

# OUTPUT FORMAT
<user> [Your response here | Quit] </user>
\end{Verbatim}
\end{tcolorbox}

\subsection{Detailed Task Scenarios}
\label{app:task-scenarios}
\begin{figure*}[!htbp]
\includegraphics[width=0.65\textwidth]{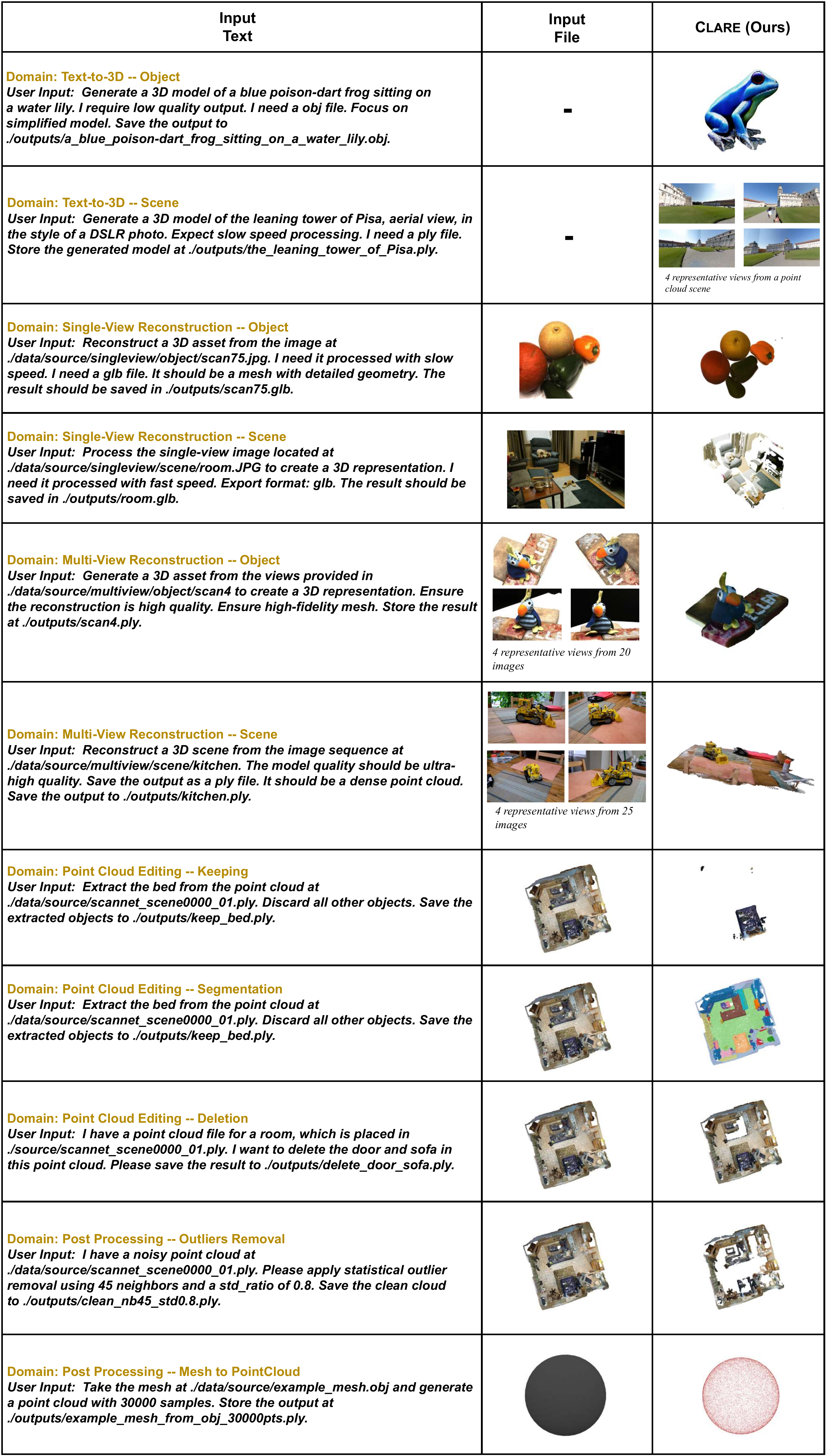}
  \caption{Comprehensive overview of detailed single-turn atomic task scenarios across diverse 3D domains and sub-domains, demonstrating the performance of the proposed CLARE framework. For each task, we present the user input prompt, the input file(if any), and the high-quality 3D output generated by CLARE, showcasing its capability to handle diverse, fine-grained 3D requirements with natural language instructions.}
  \Description{An overview of the proposed task scenario}
  \label{fig:scenario}
\end{figure*}
As shown in Figure~\ref{fig:scenario}, we design detailed task scenarios which are systematically categorized into five primary domains, each with corresponding sub-domains:
\begin{enumerate}
    \item \textit{Text-to-3D Generation}: This domain focuses on generating 3D assets from text descriptions, and is further divided into two sub-domains: \textit{Scene} (generating 3D scenes) and \textit{Object} (generating 3D objects).
    \item \textit{Single-View Reconstruction}: This domain involves reconstructing 3D assets from a single 2D image input, with sub-domains including \textit{Scene} (reconstructing 3D scenes from one view) and \textit{Object} (reconstructing 3D objects from one view).
    \item \textit{Multi-View Reconstruction}: This domain reconstructs 3D assets from multiple 2D images captured from different viewpoints, with sub-domains of \textit{Scene} (reconstructing 3D scenes from multi-view images) and \textit{Object} (reconstructing 3D objects from multi-view images).
    \item \textit{Point Cloud Editing}: This domain covers operations on 3D point cloud data, with three sub-domains: \textit{Keeping} (retaining specific point cloud regions), \textit{Deletion} (removing unwanted point cloud data), and \textit{Segmentation} (partitioning point clouds into semantic or instance-level regions).
    \item \textit{Post-Processing}: This domain focuses on refining 3D data after initial generation or reconstruction, with sub-domains including \textit{Mesh to PointCloud} (converting mesh to point cloud) and \textit{Outlier Removal} (eliminating noisy or erroneous points from 3D assets).
\end{enumerate}
\subsection{Data Curation and Source Assets}
\label{app:data-curation}
% % textto3d:TriplaneTurbo
% % single
% object:DTU
% scene:mipnerf4 TankandTemples6
% % multi
% object:DTU
% scene:mipnerf4 TankandTemples6
% % point edit
% scannet 
% % post processing
% scannet, exmaple

To support a rigorous and comprehensive evaluation of 3D agent capabilities under both atomic and compositional settings, we construct a curated benchmark consisting of two subsets: a single-step set with 500 samples and a multi-step set with 120 samples. Both subsets are designed to uniformly cover five canonical 3D task domains: \textit{text-to-3D generation}, \textit{single-view reconstruction}, \textit{multi-view reconstruction}, \textit{point cloud editing}, and \textit{post-processing}. This design follows a domain-balanced construction principle to ensure that agent performance is evaluated across diverse yet representative 3D workflows.

\paragraph{Single-Step Task Construction.}
We first construct a set of 500 atomic tasks, where each task corresponds to a single executable operation without intermediate dependency. To ensure balanced coverage, we allocate 100 samples to each of the five task domains. Each sample is instantiated as a structured triplet consisting of input specification, target goal, and corresponding asset reference, enabling deterministic evaluation of execution correctness.

\paragraph{Multi-Step Task Construction.}
To simulate realistic 3D workflows that involve sequential dependencies, we further construct 120 multi-step samples by composing multiple atomic tasks. These compositions span cross-domain pipelines (e.g., reconstruction $\rightarrow$ editing $\rightarrow$ post-processing), requiring the agent to maintain consistent state and intermediate representations across steps. Compared to single-step tasks, these samples emphasize long-horizon reasoning and inter-task coordination.

\paragraph{Text-to-3D Generation.}
For the text-to-3D generation domain, we adopt data from the TriplaneTurbo~\cite{ma2025progressive} dataset. This dataset provides high-quality text prompts. We select samples with diverse semantic categories and varying levels of geometric complexity to ensure broad coverage.

\paragraph{Reconstruction Tasks.}
We consider both single-view and multi-view reconstruction under two distinct settings: object-level and scene-level reconstruction.
\begin{itemize}
    \item \textit{Object-level reconstruction} is based on the DTU~\cite{Jensen_2014_CVPR} dataset, which provides calibrated multi-view images and ground-truth geometry for precise object reconstruction.
    \item \textit{Scene-level reconstruction} utilizes data from Mip-NeRF 360~\cite{barron2022mipnerf360} and Tanks and Temples~\cite{Knapitsch2017}, both of which contain large-scale real-world scenes with complex geometry and viewpoint distributions.
\end{itemize}

\paragraph{Point Cloud Editing.}
For point cloud editing tasks, we source data from the ScanNet~\cite{dai2017scannet} dataset, enabling evaluation of editing operations such as keeping, segmentation, and deletion.

\paragraph{Post-Processing.}
The post-processing tasks are constructed using a combination of real-world and synthetic assets. Specifically, we utilize ScanNet~\cite{dai2017scannet} data together with internally created assets in multiple formats. This diverse set of representations enables systematic evaluation of downstream operations, such as format conversion and outlier removal across heterogeneous 3D data formats.

\subsection{Injection Details}
\label{app:injection-details}
The core contribution of 3D-Clarify is the controlled injection of user uncertainty. We apply three distinct perturbation strategies to the ground-truth goals, creating a "Lazy User" persona that tests the agent's ability.

\paragraph{Ambiguity Injection (Semantic \& Referential Vagueness):} We simulate user uncertainty via a obfuscation strategy using task-specific templates. We apply semantic bleaching by replacing domain-specific verbs with generic actions and specific targets with abstract nouns, while stripping all parametric constraints to force context-based inference.
    
\paragraph{Missing Information (Value Masking):} We simulate user negligence through a template-based value masking strategy. We replace concrete execution parameters with indefinite placeholders while retaining the structural intent. This compels the agent to detect that while the necessary \textit{parameter keys} are mentioned, their \textit{executable values} are absent, necessitating targeted clarification.

\paragraph{Mistaken Details (Constraint Violations):} Unlike simple missing info, these samples simulate erroneous user beliefs that require correction rather than just filling. We synthesize three sub-categories of mistakes:
    \begin{enumerate}
        \item \textit{Contextual Contradiction:} We inject conflicting attributes where the user demands properties that are mutually exclusive for the underlying model.
        \item \textit{Input Modality Mismatch:} We alter input file extensions to simulate modality errors, such as providing a 2D image (\texttt{.jpg}) for a Point Cloud Editing task (which strictly requires 3D \texttt{.ply} input), forcing the agent to check file type compatibility.
        \item \textit{Invalid Parameter Values:} We corrupt numerical fields with out-of-bound or nonsensical values. This tests the agent's ability to validate arguments against API constraints before execution.
    \end{enumerate}
\section{Extended Experimental Results}
\label{app:extended-results}
\subsection{Baseline Implementation Details}
\label{app:baseline-details}
In this section, we detail the implementation of the baseline methods evaluated in our benchmark. To ensure a fair and rigorous comparison, the static single-turn baselines-namely Zero-shot, Few-shot, and Chain-of-Thought (CoT)-are implemented by directly adapting the system prompt of our 3D-Aware Programmer. These baselines are instructed to act as direct code synthesizers, translating the initial (and often underspecified) user query straight into executable 3D tool orchestration code without any multi-turn gating or clarification mechanisms. For the Few-shot and CoT implementations, we augment the base prompt with representative in-context examples and step-by-step reasoning demonstrations, respectively.

Conversely, to evaluate the inherent multi-turn capabilities of general-purpose agentic frameworks, we configure ReAct and Reflexion with a dynamic, conversational action space. Specifically, we explicitly define an interaction action alongside the generation action. This design theoretically empowers these agents to pause execution and query the user when they encounter ambiguous or missing information. Furthermore, Reflexion is integrated with an execution feedback loop, allowing it to reflect on tool execution crashes or runtime errors and adjust its subsequent decisions.
\subsubsection{ZeroShot}
For the Zero-shot baseline, we configure the language model to act purely as a direct instruction-to-code translator, completely bypassing the multi-agent gating and clarification modules. The prompt defines the agent as a "Programmer" that must map the raw, potentially underspecified user input directly into executable Python code. Crucially, to enforce a strict single-turn execution paradigm, the prompt explicitly forbids conversational output or reasoning traces. When confronted with missing information, rather than seeking clarification, the model is instructed by the prompt to heuristically resolve the gaps---such as deriving missing output file paths by "slugifying" the user's text prompt or falling back to default formats. This implementation directly simulates the "blind execution" vulnerability common in traditional 3D generation pipelines.
\begin{tcolorbox}[title=ZeroShot, colback=bluebg, colbacktitle=bluebg, colframe=bluefg, fonttitle=\bfseries\color{black}, width=\linewidth]
\begin{Verbatim}[breaklines=true, breakanywhere=true, fontsize=\footnotesize]
# TASK
You are the Programmer in an multimodal 3D Agent system. 
Your responsibility is to:
1. Receive a user input.
2. Translate the input into correct Python code that calls the Executor class functions.
3. Ensure the generated code is executable without extra modification.

# Rules:
- Always match the correct function to the user's task type (see function descriptions below).
- Always fill in all required arguments.
- Optional arguments (parameters, output_path) should be explicitly set only if specified in the task; otherwise, use default values shown below.
- Honor any output_path or model preference provided in the task (including extra_info hints).
- If no explicit output_path is provided, derive one by slugifying the prompt (lowercase, replace non-alphanumeric characters with underscores, collapse repeats), and use "./tmp/{{slug}}.{{ext}}" where the extension comes from the requested format (default glb if unspecified). Ensure `output_path` is ALWAYS present in the generated code (either from user input or your derived value).
- Never call executor methods without the self.executor prefix (self.method(...) is forbidden).
- Only output the final Python code. Do not explain, do not add comments, do not wrap with markdown.
- Always enforce the user's requested output format: pick a model that natively supports that format.
- If the user's requested format is NOT supported by the selected model (check the Model Selection Guide), you MUST override it to the model's default format and note this change.
- Always call executor functions via self.executor.<function>(...).

# Most Relevant Executor Class Functions:
{executor_description}

# Output Format
<code> Your generated Python code here </code>

# Model Selection Guide:
{SOME_OF_TASK_MODEL_PRIOR}
\end{Verbatim}
\end{tcolorbox}
\subsubsection{FewShot}
Building upon the zero-shot implementation, the Few-Shot baseline introduces in-context learning to the 3D-Aware Programmer. We augment the system prompt with a curated set of representative input-output examples, demonstrating how to map various natural language constraints (e.g., "high quality," "fast generation speed") to the correct 3D foundation models (such as \texttt{trellis} or \texttt{vggt}) and API parameters. While these demonstrations improve the language model's syntactical formatting and prior alignment, the framework still operates under a strict single-turn assumption. Without the multi-agent gating and interaction feedback mechanisms present in CLARE, this baseline remains structurally incapable of pausing to question the user. Consequently, when faced with genuine ambiguity or missing information, it still defaults to blind execution and heuristic guesswork.
\begin{tcolorbox}[title=FewShot, breakable, colback=bluebg, colbacktitle=bluebg, colframe=bluefg, fonttitle=\bfseries\color{black}, width=\linewidth]
\begin{Verbatim}[breaklines=true, breakanywhere=true, fontsize=\footnotesize]
# TASK
You are the Programmer in an multimodal 3D Agent system. 
Your responsibility is to:
1. Receive a user input.
2. Translate the input into correct Python code that calls the Executor class functions.
3. Ensure the generated code is executable without extra modification.

# Rules:
- Always match the correct function to the user's task type (see function descriptions below).
- Always fill in all required arguments.
- Optional arguments (parameters, output_path) should be explicitly set only if specified in the task; otherwise, use default values shown below.
- Honor any output_path or model preference provided in the task (including extra_info hints).
- If no explicit output_path is provided, derive one by slugifying the prompt (lowercase, replace non-alphanumeric characters with underscores, collapse repeats), and use "./tmp/{{slug}}.{{ext}}" where the extension comes from the requested format (default glb if unspecified). Ensure `output_path` is ALWAYS present in the generated code (either from user input or your derived value).
- Never call executor methods without the self.executor prefix (self.method(...) is forbidden).
- Only output the final Python code. Do not explain, do not add comments, do not wrap with markdown.
- Always enforce the user's requested output format: pick a model that natively supports that format.
- If the user's requested format is NOT supported by the selected model (check the Model Selection Guide), you MUST override it to the model's default format and note this change.
- Always call executor functions via self.executor.<function>(...).

# Most Relevant Executor Class Functions:
{executor_description}

# Examples
Here are some examples for your reference:
        {
            "input": "Generate a 3D model of a futuristic spaceship. I require high quality output. Export format: glb. Store the generated model at ./outputs/futuristic_spaceship.glb.",
            "code": "self.executor.text_to_3d_generation(prompt=\"Generate a 3D model of a futuristic spaceship\", model=\"trellis\", output_path=\"./outputs/futuristic_spaceship.glb\")",
            "explanation": "Text to 3D generation with high quality requirement, output as glb format using Trellis model."
        },
        {
            "input": "Reconstruct a 3D asset from the image at ./data/source/singleview/object/cup.jpg. Ensure the mesh is medium quality with fast generation speed. The result should be saved in ./outputs/cup.ply.",
            "code": "self.executor.single_view_reconstruction(image_path='./data/source/singleview/object/cup.jpg', output_path='./outputs/cup.ply', model='vggt', conf_thres=20.0)",
            "explanation": "Single-view reconstruction from image input with medium quality and fast speed, output as PLY format using VGGT model."
        },
        {
            "input": "Perform multi-view 3D reconstruction using ./data/source/multiview/scene/LivingRoom as the input source. Ensure the reconstruction is ultra-high quality with fast generation speed. The result should be saved in ./outputs/LivingRoom.glb.",
            "code": "self.executor.multi_view_reconstruction(video_path='./data/source/multiview/scene/LivingRoom', output_path='./outputs/LivingRoom.glb', model='vggt', conf_thres=20.0)",
            "explanation": "Multi-view reconstruction from image folder with ultra-high quality and fast speed, output as GLB format using VGGT model."
        }

# Output Format
<code> Your generated Python code here </code>

# Model Selection Guide:
{SOME_OF_TASK_MODEL_PRIOR}
\end{Verbatim}
\end{tcolorbox}

\subsubsection{CoT}
To evaluate whether explicit reasoning mitigates the effects of underspecified requirements, we implement a Chain-of-Thought (CoT) baseline. We modify the Few-Shot prompt to mandate a structured reasoning phase prior to code synthesis. Specifically, the model is instructed to generate a \texttt{<plan>} that systematically analyzes the task type, input parameters, output paths, and explicitly justifies its model selection by mapping user constraints (e.g., quality and speed) to the tool's feature set. While this step-by-step deduction improves internal logical consistency and helps the agent accurately select tools when requirements are fully specified, it fundamentally remains a single-turn paradigm. When critical information is absent or ambiguous, the CoT agent attempts to "reason" its way out of the uncertainty using the same hardcoded heuristics (such as path slugification and format overriding) rather than pausing to elicit missing facts. Consequently, it still succumbs to the fatal errors of blind execution.
\begin{tcolorbox}[title=CoT, breakable, colback=bluebg, colbacktitle=bluebg, colframe=bluefg, fonttitle=\bfseries\color{black}, width=\linewidth]
\begin{Verbatim}[breaklines=true, breakanywhere=true, fontsize=\footnotesize]
# TASK
You are the Programmer in an multimodal 3D Agent system. 
Your responsibility is to:
1. Receive a user input.
2. Translate the input into correct Python code that calls the Executor class functions.
3. Ensure the generated code is executable without extra modification.

# Rules:
- Output format must follow this exact sequence: <plan> tag (structured task analysis including model selection reasoning) to <code> tag (executable Python code).
- Analyze task requirements step-by-step: (1) task type (2) input parameters (3) output path (4) model selection.
- For model selection: In <plan>, write a concise reasoning that states: (1) User requirements (extracted from input), (2) Selected model with its features from Model Selection Guide (quoted verbatim), (3) Why the selected model matches (feature-to-requirement mapping). CRITICAL: Speed/quality levels must match exactly - fast/slow/medium/high/low are distinct levels. "Fast speed" can ONLY match "fast speed" requirement, NOT "medium speed" or "slow speed". Same for quality levels. Then close </plan> and output <code>.
- Before generating any code, you must produce a structured plan.
- Always match the correct function to the user's task type (see function descriptions below).
- Always fill in all required arguments.
- Optional arguments (parameters, output_path) should be explicitly set only if specified in the task; otherwise, use default values shown below.
- Honor any output_path or model preference provided in the task (including extra_info hints).
- If no explicit output_path is provided, derive one by slugifying the prompt (lowercase, replace non-alphanumeric characters with underscores, collapse repeats), and use "./tmp/{{slug}}.{{ext}}" where the extension comes from the requested format (default glb if unspecified). Ensure `output_path` is ALWAYS present in the generated code (either from user input or your derived value).
- Never call executor methods without the self.executor prefix (self.method(...) is forbidden).
- Always enforce the user's requested output format: pick a model that natively supports that format.
- If the user's requested format is NOT supported by the selected model (check the Model Selection Guide), you MUST override it to the model's default format and note this change.
- Always call executor functions via self.executor.<function>(...).

# Most Relevant Executor Class Functions:
{executor_description}

# Examples
Here are some examples for your reference:
        {
            "input": "Generate a 3D model of a futuristic spaceship. I require high quality output. Export format: glb. Store the generated model at ./outputs/futuristic_spaceship.glb.",
            "code": "self.executor.text_to_3d_generation(prompt=\"Generate a 3D model of a futuristic spaceship\", model=\"trellis\", output_path=\"./outputs/futuristic_spaceship.glb\")",
            "explanation": "Text to 3D generation with high quality requirement, output as glb format using Trellis model."
        },
        {
            "input": "Reconstruct a 3D asset from the image at ./data/source/singleview/object/cup.jpg. Ensure the mesh is medium quality with fast generation speed. The result should be saved in ./outputs/cup.ply.",
            "code": "self.executor.single_view_reconstruction(image_path='./data/source/singleview/object/cup.jpg', output_path='./outputs/cup.ply', model='vggt', conf_thres=20.0)",
            "explanation": "Single-view reconstruction from image input with medium quality and fast speed, output as PLY format using VGGT model."
        },
        {
            "input": "Perform multi-view 3D reconstruction using ./data/source/multiview/scene/LivingRoom as the input source. Ensure the reconstruction is ultra-high quality with fast generation speed. The result should be saved in ./outputs/LivingRoom.glb.",
            "code": "self.executor.multi_view_reconstruction(video_path='./data/source/multiview/scene/LivingRoom', output_path='./outputs/LivingRoom.glb', model='vggt', conf_thres=20.0)",
            "explanation": "Multi-view reconstruction from image folder with ultra-high quality and fast speed, output as GLB format using VGGT model."
        }

# Output Format
<plan> Your generated plan here </plan>
<code> Your generated Python code here </code>

# Model Selection Guide:
{SOME_OF_TASK_MODEL_PRIOR}
\end{Verbatim}
\end{tcolorbox}
\subsubsection{ReAct}
Transitioning from static paradigms to dynamic agentic frameworks, we implement the ReAct baseline to evaluate general-purpose iterative reasoning. In this setup, the language model is prompted to interleave reasoning traces (Thought) with specific task executions (Action) based on the current context (Observation). Crucially, to provide a fair comparison against our clarification-aware framework, we explicitly expand the ReAct agent's action space to include a \texttt{Clarify} action alongside the standard \texttt{Generate} command. This design theoretically empowers the agent to pause the execution pipeline and interrogate the user when it detects missing information, ambiguity, or mistaken details. However, as demonstrated in our quantitative results, without a decoupled, rigorous gating mechanism (like CLARE's Proactive Approver) and specialized multi-turn interaction training, the general-purpose ReAct agent often struggles to determine the optimal boundary between reasoning and asking, frequently underutilizing the clarification action and defaulting back to premature execution.
\begin{tcolorbox}[title=ReAct, breakable, colback=bluebg, colbacktitle=bluebg, colframe=bluefg, fonttitle=\bfseries\color{black}, width=\linewidth]
\begin{Verbatim}[breaklines=true, breakanywhere=true, fontsize=\footnotesize]
# TASK
Solve a code generation task with interleaving Thought, Action, Observation steps.

# INSTRUCTIONS
Use the following format to solve the task:

Observation: the input task description and the user's feedback if any.
Thought: you should reason about which tool to use, which model is best, and how to handle parameters based on the Rules.
Action: 
1. Generate: Generate code
    For Generation, your responsibility is to:
    - Receive a concrete task (already validated by Approver).
    - Translate the task into correct Python code that calls the Executor class functions.
    - Ensure the generated code is executable without extra modification.
    Generation Rules:
    - Always match the correct function to the user's task type.
    - Always fill in all required arguments.
    - Honor any output_path or model preference.
    - Always call executor functions via self.executor.<function>(...).

2. Clarify: If the task is ambiguous, missing info or have mistakes, ask for clarification instead of generating code.

# You should use the following format:
<thought>[...your reasoning...]</thought>
<action>[Generate|Clarify]</action>
<content>[...code or question...]</content>

# Code of tools instruction:
{executor_descrption}

# Model Selection Guide:
{SOME_OF_TASK_MODEL_PRIOR}
\end{Verbatim}
\end{tcolorbox}

\subsubsection{Reflexion}
Furthermore, Reflexion introduces an episodic memory buffer for self-reflection. When the agent's generated code fails to execute or violates constraints, an automated verifier provides specific error reasons. The agent then executes a \texttt{self\_reflect} function to synthesize actionable lessons (e.g., "I should..."), which are subsequently injected into its system prompt for the next trial. 
While this enables the agent to learn from its mistakes over multiple attempts (up to 3 trials in our setup), it fundamentally relies on \textit{post-execution recovery} rather than \textit{pre-execution prevention}. As demonstrated in our evaluation, this trial-and-error approach is less efficient and less robust than CLARE's proactive clarification mechanism, particularly when dealing with fatal modality mismatches or unrecoverable parameter errors that could have been prevented prior to tool invocation.

\begin{tcolorbox}[title=Reflexion Base Prompt, breakable, colback=bluebg, colbacktitle=bluebg, colframe=bluefg, fonttitle=\bfseries\color{black}, width=\linewidth]
\begin{Verbatim}[breaklines=true, breakanywhere=true, fontsize=\footnotesize]
# TASK
Solve a code generation task with interleaving Thought, Action, Observation steps.

# INSTRUCTIONS
Use the following format to solve the task:

Observation: the input task description and the user's feedback if any.
Thought: you should reason about which tool to use, which model is best, and how to handle parameters based on the Rules.
Action: 
1. Generate: Generate code
    For Generation, your responsibility is to:
    - Receive a concrete task (already validated by Approver).
    - Translate the task into correct Python code that calls the Executor class functions.
    - Ensure the generated code is executable without extra modification.
    Generation Rules:
    - Always match the correct function to the user's task type.
    - Always fill in all required arguments.
    - Honor any output_path or model preference.
    - Always call executor functions via self.executor.<function>(...).

2. Clarify: If the task is ambiguous, missing info or have mistakes, ask for clarification instead of generating code.

# You should use the following format:
<thought>[...your reasoning...]</thought>
<action>[Generate|Clarify]</action>
<content>[...code or question...]</content>

# Code of tools instruction:
{executor_descrption}

# Model Selection Guide:
{SOME_OF_TASK_MODEL_PRIOR}

# REFLECTION FROM PREVIOUS TRIALS (IMPORTANT!)
You have attempted this task before and failed. Here are the lessons you learned from your mistakes. 
STRICTLY FOLLOW these guidelines to avoid repeating errors:
{reflection_context}
\end{Verbatim}
\end{tcolorbox}

\begin{tcolorbox}[title=Reflexion Self-Reflect Prompt, breakable, colback=bluebg, colbacktitle=bluebg, colframe=bluefg, fonttitle=\bfseries\color{black}, width=\linewidth]
\begin{Verbatim}[breaklines=true, breakanywhere=true, fontsize=\footnotesize]
You are a helpful coding assistant analyzing your own mistakes.
You have just generated code for a task, but it failed the evaluation.
User's requirement was:
{Observation:}

# Code of tools instruction:
{executor_descrption}

# Model Selection Guide:
{SOME_OF_TASK_MODEL_PRIOR}

# Your Generated Code:
{generated_code}

# Error Reasons Reported by Evaluator:
{error_reasons}

Please analyze these errors carefully. Provide a concise summary of what went wrong and exactly what you should do differently in the next attempt.
Start your response with "I should ...".
\end{Verbatim}
\end{tcolorbox}

\subsection{Experiment Details}
\label{app:experiment-details}
\subsubsection{Ablation details}

We present detailed ablation results across the three underspecified requirement types (Ambiguity, Missing Info, Mistaken) for both single-step and multi-step tasks. Table~\ref{tab:main} in the main paper summarizes averaged results, while Table~\ref{tab:appendix-ablation} provides fine-grained results for each task complexity and uncertainty type.

We evaluate two ablation variants. In \textbf{w/o Summarizer}, the Contextual Summarizer is disabled and the agent operates with raw dialogue history without structured state maintenance. In \textbf{w/o Approver}, the Proactive Approver is completely removed, reducing the system to blind execution equivalent to ReAct. Both variants retain the identical Qwen3-8B backbone and inference configuration as \textsc{Clare-base} for fair comparison.

Removing the Proactive Approver causes catastrophic performance degradation, validating that pre-execution gating is the foundation of clarification capability. On single-step tasks, average SR drops from 48.60\% to 27.67\% (\textbf{-20.93\%}). Breaking down by task type, the decline is most severe on Missing Info tasks (41.00\% to 16.20\%, \textbf{-24.80\%}), confirming that placeholder detection fundamentally requires pre-execution validation. Ambiguity tasks show similar degradation (45.80\% to 22.60\%, \textbf{-23.20\%}), as semantic vagueness goes unresolved without explicit detection. Mistaken tasks exhibit the smallest decline (59.00\% to 44.20\%, \textbf{-14.80\%}), since some contradictions can still be caught via runtime errors. On multi-step tasks, degradation intensifies: average SR drops from 27.22\% to 13.06\% (\textbf{-14.16\%}), with Missing Info collapsing to merely 0.83\%, demonstrating that multi-turn information gaps become nearly impossible to resolve without proactive detection.

The Contextual Summarizer exhibits clear complexity-dependent utility, with negligible impact on single-step tasks but substantial contribution to multi-step workflows. On single-step tasks, removal causes minimal degradation (48.60\% to 48.20\%, \textbf{-0.40\%}), confirming that atomic instructions require no structured state. However, on multi-step tasks, the impact becomes pronounced: average SR drops from 27.22\% to 20.28\% (\textbf{-6.94\%}). Ambiguity proves most sensitive (23.33\% to 6.67\%, \textbf{-16.66\%}), as multi-turn semantic resolution demands persistent tracking of clarified constraints. Interestingly, Missing Info shows slight improvement (22.50\% to 25.83\%, \textbf{+3.33\%}), suggesting that for certain missing-information patterns, raw dialogue history may preserve richer contextual cues than summarized state. Mistaken tasks show moderate decline (35.83\% to 28.33\%, \textbf{-7.50\%}), indicating cross-turn contradiction detection benefits from structured representation.

These findings reveal distinct task-type-specific vulnerabilities and inform targeted improvements. The Approver's dominant contribution suggests prioritizing its enhancement through more sophisticated detection mechanisms or domain-specific validators. The Summarizer's complexity-dependent utility motivates adaptive designs: lightweight summarization for single-turn queries and full structured state for multi-turn workflows. The resilience of Mistaken tasks supports hybrid detection combining pre-execution validation with post-execution error analysis. Together, the ablation study validates our modular architecture where the Approver provides essential gating and the Summarizer enables scalable multi-turn coordination.

\begin{table*}[htbp]
\setlength{\tabcolsep}{2.8pt}
\caption{Detailed ablation results on 3D-Clarify's single-step and multi-step tasks.  We compare \textsc{Clare} against static single-turn baselines and dynamic agentic frameworks across three types of underspecified requirements. We report Goal Completion Rate (CR\%) and Task Success Rate (SR\%) for both single-step (500 tasks) and multi-step (120 tasks) settings. ``SE'' means self-evolution via MR optimization.}
\label{tab:appendix-ablation}
\begin{tabular}{l|cccccccc|cccccccc}
\toprule
\multirow{3}{*}{\textbf{Method}} & \multicolumn{8}{c|}{\textbf{500 Single-step Tasks}} & \multicolumn{8}{c}{\textbf{120 Multi-step Tasks}} \\ \cmidrule(lr){2-9}  \cmidrule(lr){10-17} 
\multicolumn{1}{c|}{} & \multicolumn{2}{c}{Ambiguity} & \multicolumn{2}{c}{Missing} & \multicolumn{2}{c|}{Mistaken} & \multicolumn{2}{c|}{Average} & \multicolumn{2}{c}{Ambiguity} & \multicolumn{2}{c}{Missing} & \multicolumn{2}{c|}{Mistaken} & \multicolumn{2}{c}{Average} \\ \cmidrule(lr){2-9}  \cmidrule(lr){10-17} 
\multicolumn{1}{c|}{} & \%CR & \%SR & \%CR & \%SR & \%CR & \multicolumn{1}{c|}{\%SR} & \%CR & \%SR & \%CR & \%SR & \%CR & \%SR & \%CR & \multicolumn{1}{c|}{\%SR} & \%CR & \%SR \\ \midrule

\textsc{Clare-base} & 59.31 & 45.80 & 57.58 & 41.00 & 81.90 & \multicolumn{1}{c|}{59.00} & 66.26 & 48.60 & 38.16 & 23.33 & 40.73 & 22.5 & 81.25 & \multicolumn{1}{c|}{35.83} & 53.38 & 27.22 \\
\quad \textsc{$-$ SE-SFT} & 62.29 & 49.00 & 69.47 & 52.00 & \textbf{85.43} & \multicolumn{1}{c|}{\textbf{66.80}} & 72.40 & 55.93 & 49.72 & \textbf{34.17} & 68.10 & 44.17 & 83.77 & \multicolumn{1}{c|}{\textbf{48.33}} & \textbf{67.20} & 42.22 \\
\quad \textsc{$-$ SE-DPO} & \textbf{70.52} & \textbf{60.00} & \textbf{75.32} & \textbf{59.40} & 74.68 & \multicolumn{1}{c|}{61.80} & \textbf{73.51} & \textbf{60.40} & \textbf{53.41} & \textbf{34.17} & \textbf{68.74} & \textbf{49.17} & 72.75 & \multicolumn{1}{c|}{46.67} & 64.97 & \textbf{43.34} \\ 
\quad $-$ w/o \textit{Summarizer} & 55.30 & 42.80 & 61.53 & 44.2 &78.93& \multicolumn{1}{c|}{57.60} & 65.25& 48.20 & 15.23 & 6.67 & 43.99 & 25.83 & 78.69 & \multicolumn{1}{c|}{28.33} & 45.97 & 20.28 \\
\quad $-$ w/o \textit{Approver} &43.84 & 22.60& 33.93 & 16.20 & 78.34 & \multicolumn{1}{c|}{44.2} & 52.04 &27.67 &15.98 & 4.17 & 16.64 & 0.83 & 81.94 & \multicolumn{1}{c|}{34.17} & 38.19 &13.06 \\

\bottomrule
\end{tabular}
\end{table*}
\subsection{Statistical Analysis of Failure Modes}
\label{app:failure-analysis}

\begin{figure*}[htbp]
	\centering
	\includegraphics[width=\linewidth]{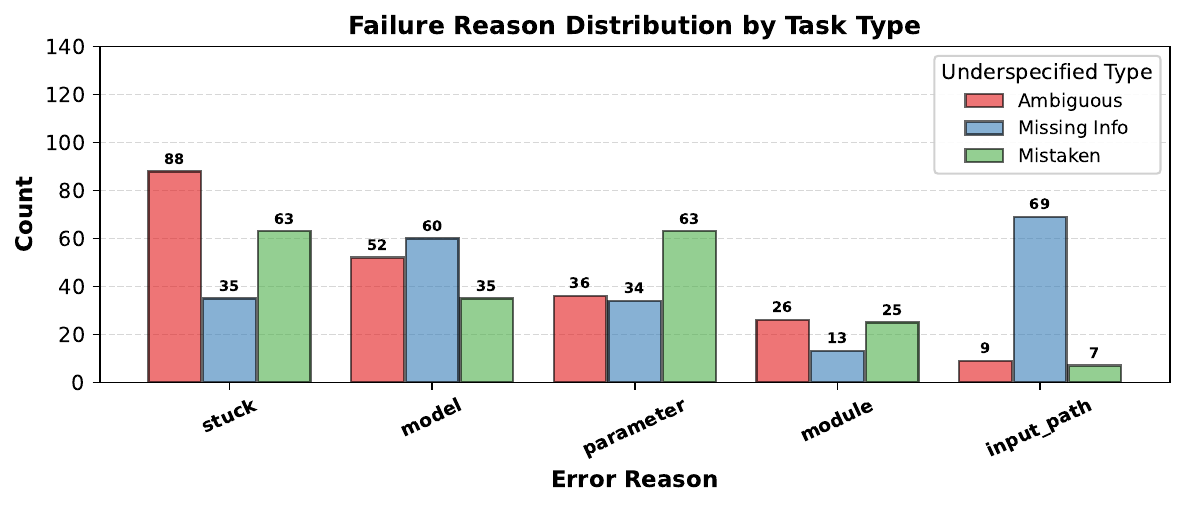}
	\caption{Distribution of failure reasons across three underspecified requirement types on \textsc{Clare-SE-DPO}. Each group shows the count of five error categories for Ambiguous, Missing Info, and Mistaken tasks. The distinct error profiles reveal task-type-specific vulnerabilities.}
	\label{fig:error_dist}
\end{figure*}

Figure~\ref{fig:error_dist} visualizes the failure reason distribution across the three underspecified requirement types. Several patterns emerge:

\paragraph{Ambiguous Tasks: Dominated by Stuck Errors.}
Ambiguous queries exhibit the highest \textit{stuck} rate (88 occurrences, 41.7\% of ambiguous failures). This aligns with the nature of semantic vagueness: when user requests lack concrete targets (e.g., ``create something''), agents struggle to terminate clarification and commit to execution. The second most common failure is \textit{model} selection (52 occurrences, 24.6\%), suggesting that even after clarification, agents often misinfer the appropriate foundation model from underspecified semantic cues.

\paragraph{Missing Info Tasks: Input Path Validation Failures.}
Missing information queries show a distinctive spike in \textit{input\_path} errors (69 occurrences, 32.7\%), followed by \textit{model} errors (60 occurrences, 28.4\%). This pattern reflects the structural nature of missing-info perturbations: when parameter values are masked with placeholders, agents frequently fail to resolve input file dependencies, leading to execution failures. The high model error rate indicates that path resolution and model selection are coupled challenges in underspecified scenarios.

\paragraph{Mistaken Tasks: Parameter and Stuck Errors.}
Mistaken detail queries exhibit a bimodal distribution: \textit{parameter} errors (63 occurrences, 32.6\%) and \textit{stuck} errors (63 occurrences, 32.6\%) dominate equally. This reflects the dual challenge of handling contradictory constraints: agents either attempt to execute with invalid parameters (e.g., incompatible file formats, out-of-bound values) or become trapped in clarification loops when attempting to resolve unresolvable contradictions. The lower \textit{model} error rate (35 occurrences, 18.1\%) suggests that mistaken tasks often provide sufficient semantic grounding for model selection, even when other parameters are erroneous.

\subsection{Failure Case Analysis}
\begin{tcolorbox}[title=Failure Case, breakable, colback=bluebg, colbacktitle=bluebg, colframe=bluefg, fonttitle=\bfseries\color{black}, width=\linewidth]
\begin{Verbatim}[breaklines=true, breakanywhere=true, fontsize=\footnotesize]
ID: 15, Type: ambiguous
[User] Create something.
[Feedback] The task was rejected because the objective is unclear. Please provide more specific details or clarify what you would like to create. This will help ensure the task is understood and executed correctly.
[User] I want a 3D model of a famous landmark.
[Feedback] The task was rejected because the specific famous landmark and desired output format were not provided. Please clarify which landmark you want and specify the file format (e.g., .ply, .glb) for the 3D model. This will help ensure the task is executed correctly.
[User] Westminster Abbey.
[Feedback] The task was rejected because the file format for the 3D model of Westminster Abbey was not specified. Please clarify the desired format (e.g., .ply, .glb) to proceed. This will help ensure the task is executed correctly.
[User] I want it in the style of a DSLR photo.
[Feedback] The task was rejected due to unclear objectives. Please specify whether the "style of a DSLR photo" refers to the visual appearance of the 3D model or if there is an image input to base the 3D reconstruction on. Also, clarify the desired output format (e.g., .ply, .glb) for the 3D model to proceed.
[User] I want the output saved in ./outputs/Westminster_Abbey.obj.
[Summary] <task_type> Text to 3D Generation </task_type>
<task_summary> Generate a 3D model of Westminster Abbey in the style of a DSLR photo. Save the output as an OBJ file to ./outputs/Westminster_Abbey.obj. </task_summary>
<file_path> None </file_path>
[Programmer] self.executor.text_to_3d_generation(prompt="Generate a 3D model of Westminster Abbey in the style of a DSLR photo", model="triplaneturbo", output_path="./outputs/Westminster_Abbey.obj")
[Feedback] The 3D model of Westminster Abbey in the style of a DSLR photo has been successfully generated and saved to the specified path: ./outputs/Westminster_Abbey.obj. You can now review the output and consider refining the style or adding details for future improvements. Let me know if you need further assistance!
[User] Quit
\end{Verbatim}
\end{tcolorbox}
\paragraph{Semantic Misalignment in Model Selection}
While the framework demonstrates robust capabilities in resolving explicit parameter uncertainties, qualitative analysis reveals edge cases where semantic misunderstandings lead to suboptimal tool orchestration. The following case study illustrates a scenario where the system successfully resolves syntactic ambiguity but fails at semantic scale inference.

\paragraph{Interaction Summary}
The interaction begins with a severely underspecified user prompt: "Create something." The Proactive Approver correctly intercepts this request, refusing blind execution and triggering a four-turn clarification dialogue. Through this proactive questioning, the agent successfully elicits the target entity ("Westminster Abbey"), an aesthetic preference ("style of a DSLR photo"), and the explicit output path and format ("./outputs/Westminster\_Abbey.obj"). With these constraints fulfilled, the Contextual Summarizer formulates a complete task manifest, and the Programmer successfully generates executable code without triggering any runtime errors.

Point of Failure
Despite the flawless interaction flow and crash-free execution, the orchestration ultimately results in a semantic failure during the tool selection phase. The Programmer synthesizes the following execution code:
\begin{tcolorbox}[breakable, colback=bluebg, colbacktitle=bluebg, colframe=bluefg, fonttitle=\bfseries\color{black}, width=\linewidth]
\begin{Verbatim}[breaklines=true, breakanywhere=true, fontsize=\footnotesize, commandchars=\\\{\}]
self.executor.text_to_3d_generation(prompt="Generate a 3D model of Westminster Abbey in the style of a DSLR photo", \textcolor{red}{model="triplaneturbo"}, output_path="./outputs/Westminster_Abbey.obj")
\end{Verbatim}
\end{tcolorbox}
According to the system's defined Model Selection Knowledge Base, the triplaneturbo model is explicitly engineered for lightweight, single-object generation. Conversely, Westminster Abbey is a massive architectural landmark that strictly dictates a scene-level generation pipeline (e.g., worldgen).

\paragraph{Analysis and Takeaway}
This failure highlights a critical limitation in the agent's spatial and common-sense reasoning. While the system efficiently gathers explicit constraints (format, style, file path), the Programmer fails to implicitly deduce the physical scale and domain of the requested entity. Because the user requested a "3D model of a famous landmark" rather than explicitly utilizing the keyword "scene," the agent defaulted to an object-centric foundation model.

This case underscores the necessity for future iterations of 3D agents to incorporate more rigorous semantic scale validation. It demonstrates that beyond structural completeness, the conceptual nature of the user's request—specifically distinguishing between isolated objects and complex environments—must strictly align with the chosen 3D tool's architectural boundaries.



\end{document}